\documentclass{article}%
\usepackage{iclr2027_conference,times}

\usepackage{amsmath,amsfonts,bm}

\def\figref#1{figure~\ref{#1}}
\def\Figref#1{Figure~\ref{#1}}

\def\eqref#1{equation~\ref{#1}}
\def\Eqref#1{Equation~\ref{#1}}

\def\1{\bm{1}}

\def\vtheta{{\bm{\theta}}}

\def\vf{{\bm{f}}}

\def\vv{{\bm{v}}}

\def\vx{{\bm{x}}}

\DeclareMathAlphabet{\mathsfit}{\encodingdefault}{\sfdefault}{m}{sl}
\SetMathAlphabet{\mathsfit}{bold}{\encodingdefault}{\sfdefault}{bx}{n}

\usepackage{hyperref}
\usepackage{url}

\usepackage{graphicx}
\usepackage{enumitem}
\usepackage{cleveref}
\usepackage{xcolor}

\newcommand{\old}{\mathrm{old}}
\newcommand{\bigO}{\mathcal{O}}

\title{Elucidating the Design Space of Regression-based Diffusion Reinforcement Learning}

\author{Toyota Li, David Zhao \& Alan Zhao \\
	Tencent\\
}

\iclrfinalcopy%
\begin{document}

	\maketitle
	
	\begin{abstract}
		A nascent family of methods that forgoes the policy gradient and reweights a supervised regression instead has garnered momentum in reinforcement learning for diffusion and flow models. DiffusionNFT~\citep{zheng2026diffusionnft}, FlowAWR~\citep{2026arXiv260630376F}, and RAM~\citep{2026arXiv260510759B} are representative regimes with contrasting motivations. It is yet opaque what, if anything, they share. We substantiate that each is the solution of one divergence-constrained reward-maximization problem, and they are differentiated only by the convex generator that defines the constraint. Under the unified modeling framework, we unravel the relaxations that prior art made during building the advantage-embedded regression target: approximating the KKT condition and posterior normalizer for the linear and exponential tilt shapes DiffusionNFT and FlowAWR respectively, while preserving the exact sparsemax projection onto the probability simplex for linear tilt leads to another superior model type in this work. Beyond the theoretical underpinnings, we further empirically investigate the design space and shed light on the training recipe for regression-style diffusion RL. Retaining the merits discovered during our exploration gives rise to DiffusionRFT, our paradigm that converges faster, trains more stably, and attains the top performance.
	\end{abstract}
	
	\section{Introduction}
	
	Post-training with an external reward plays a pivotal role in endowing a diffusion model with human-preference behaviors that pre-training dictates only weakly, such as prompt adherence and text rendering. Reinforcement learning serves as a cornerstone of this utility. The trailblazers in this field are the policy gradient~\citep{NEURIPS2025_3a10c465,2025arXiv250507818X} and direct reward backpropagation~\citep{NEURIPS2023_33646ef0,clark2024directly,2024arXiv240708737P} scheme. Nevertheless, they are both cumbersome, in that a long denoising trajectory is indispensable and exhibits high variance. Moreover, the training objective diverges from the simple Mean-Squared Error (MSE) loss prevalent in pre-training, that is believed to simplify the generative distribution modeling to supervised learning, especially at scale.
	
	Recently, a growing body of forward-process policy optimization ushers in a new wave of diffusion RL, including DiffusionNFT, FlowAWR and RAM, which reserve the regular diffusion loss and inject a reward into the regression target, so that a policy update is again a least-square fit on noised samples. These algorithms are grounded in divergent perspectives, ranging from CFG-style contrastive rectification~\cite{pmlr-v267-chen25ac} to stochastic optimal control~\citep{domingo-enrich2025adjoint}. Worse still, they are separately presented as standalone and integrated systems. The missing piece is a principled account of their subtle interweaving, and which design decisions drive the performance.
	
	To bridge this gap, we uncover that one \emph{optimization step} underlying these RL algorithms converges to a common driving force: maximize reward on a frozen rollout buffer under a divergence penalty. The divergence penalty translates the primary algorithmic choice into a single convex generator $f$ (\eqref{eq:density_ratio}). The reverse-KL generator yields the exponential tilt, whose group estimator coincides with the FlowAWR objective, while the Pearson $\chi^2$ generator yields a linear tilt, that DiffusionNFT already applies by standardizing the reward within a prompt group and dividing by a heuristic constant, but without such a derivation. As stated above, permitting the realization of the unified optimization objective demands an additional approximation, estimating a posterior normalizer by its endpoint counterpart. We demonstrate it preserves the direction of every step exactly while rescaling its length by a factor with unit mean and variance bounded by the second moment of the advantage.
	
	The Pearson $\chi^2$ case turns out to be the more promising one, to which most of our research endeavor is devoted. Its exact constrained optimum is a Euclidean projection onto the probability simplex, that is the sparsemax weight, whereas the linear advantage heuristically adopted by DiffusionNFT keeps only the interior branch of the Karush--Kuhn--Tucker conditions and silently drops the constraint that relative densities stay non-negative. We articulate the condition under which the dense form is exact, observe that it is violated by the dense linear tilt on most training steps, and the training collapses afterwards. Retaining the projection repairs the collapse and achieves higher rewards. We term our resulting method \emph{Diffusion Regression Fine-tuning}, or DiffusionRFT, a notably more stable and efficient point in this design space.
	
	Beyond the scope of our derivation lie the empirical design choices. The remainder is cast into four axes: the regression variable, the reward shaping and its baseline, the anchor from which the target is displaced, and the law governing the renoising timestep, and we adjust each in isolation. The regression variable lays the groundwork for subsequent algorithmic axes. Empirical results advocate the sparsemax tilt in line with our exact derivation. Meanwhile, an analytically variance-minimizing baseline further facilitates performance stabilization and enhancement across all tilts. A frozen anchor, though supported by RAM, admits no safe operating point, regardless of the update magnitude. Decoupling the renoising timestep from the generating trajectory, which might be assumed to enrich the experience buffer, instead steers the loss toward noise levels the sampler never visits.
	
	\section{Reward Maximization under a Divergence Constraint}
	\label{sec:optimal_reward}

	\subsection{Policy Improvement on a Frozen Rollout Buffer}
	\label{subsec:method_prelims}
	
	The policy-gradient family, from pioneering PPO~\citep{2017arXiv170706347S} to the critic-free estimators of GRPO~\citep{2024arXiv240203300S}, alternates between drawing trajectories from a behavior policy $\pi^\old$ and updating the current policy $\pi_\theta$ on the resulting batch. Only the outer loop is online: within one update the sampling distribution is fixed, so the batch defines a static measure $\mathcal{D}$ and the update poses an offline question --- raise the expected reward, but not so far from $\pi^\old$ that the batch stops being informative about $\pi_\theta$. As a penalty,
	\begin{equation}
		\max_{\pi_\theta}\;\; \mathbb{E}_{s\sim\mathcal{D}}\Big[\, \mathbb{E}_{a\sim\pi_\theta(\cdot|s)}\,R(s,a) \;-\; \gamma\,D\big(\pi_\theta(a|s)\,\big\|\,\pi^\old(a|s)\big) \Big],
		\label{eq:constrained_objective}
	\end{equation}
	where $s$ is a state, $a$ an action, $R$ a scalar reward, $D$ a divergence between the two conditional laws, and $\gamma>0$ sets the exchange rate between reward and conservatism.
	
	It pays to specify $D$ through the density ratio rather than through the policy itself. Writing
	\begin{equation}
		\rho(a|s) \;=\; \frac{\pi_\theta(a|s)}{\pi^\old(a|s)},
		\quad
		D_f\big(\pi_\theta\,\|\,\pi^\old\big) \;=\; \mathbb{E}_{a\sim\pi^\old(\cdot|s)}\big[\,f\!\left(\rho(a|s)\right)\,\big],
		\quad f \text{ convex},\; f(1)=0,
		\label{eq:density_ratio}
	\end{equation}
	turns $\pi_\theta$ into a scalar reweighting of a distribution $\pi^\old$ we can already sample from, and confines the regularizer's entire effect to the generator $f$. Reverse KL is $f_{\mathrm{KL}}(\rho)=\rho\log\rho$, which we adopt here; \S\ref{sec:chi2} shows that DiffusionNFT implicitly commits to a different and equally natural generator. With $f=f_{\mathrm{KL}}$ the objective is strictly concave in $\rho$ and separates across states, so a single multiplier for $\mathbb{E}_{\pi^\old}[\rho]=1$ gives the classical solution~\citep{NEURIPS2023_a85b405e}
	\begin{equation}
		\pi^*(a|s) \;=\; \frac{1}{Z(s)}\,\pi^\old(a|s)\,\exp\!\Big(\frac{R(s,a)}{\gamma}\Big),
		\qquad
		Z(s) \;=\; \mathbb{E}_{a\sim\pi^\old(\cdot|s)}\Big[\exp\!\Big(\frac{R(s,a)}{\gamma}\Big)\Big].
		\label{eq:optimal_policy_sol}
	\end{equation}
	The optimum is $\pi^\old$ times a positive, reward-monotone factor, so mass is redistributed towards high-reward actions rather than assigned to unvisited ones, and that factor is closed-form, so knowing the rewards on the batch means knowing the optimum. Together they make the problem a candidate for regression rather than for policy-gradient estimation. Taking a state to be a prompt $c$, an action a sample $x_0$, and $\pi^\old(\cdot|c)$ the terminal law $p^\old(\cdot|c)$ of the sampler, \eqref{eq:optimal_policy_sol} becomes
	\begin{equation}
		p^*(x_0 | c) \;=\; \frac{1}{Z(c)}\; p^\old(x_0 | c)\,\exp\!\Big(\frac{R(c, x_0)}{\gamma}\Big),
		\label{eq:exp_tilt_sol}
	\end{equation}
	an \emph{exponential tilt} of $p^\old$. This is the endpoint distribution an update should aim at, and it is defined without reference to how the batch was explored, in particular no stochastic sampler is needed to make sense of it. It is not yet an objective, because a flow model is parameterized by a velocity field rather than by its endpoint density.
	
	\subsection{From a Tilted Endpoint Law to a Velocity Field}
	\label{subsec:tilt_to_velocity}
	
	Flow matching~\citep{lipman2023flow,liu2023flow} sends $p_{\text{data}}$ at $t=0$ to $\mathcal{N}(0,I)$ at $t=1$ along $x_t = (1-t)\,x_0 + t\,\epsilon$, with conditional velocity $u_t(x_t|x_0) = \frac{\epsilon-x_t}{1-t}$, and fits a network by regression onto it,
	\begin{equation}
		\mathcal{L}_{\mathrm{FM}}(\theta) = \mathbb{E}_{t,\,x_0\sim p_{\text{data}},\,\epsilon\sim\mathcal{N}(0,I)}\Big[\big\| v_\theta(x_t,c,t) - u_t(x_t|x_0) \big\|^2\Big],
		\label{eq:fm_loss}
	\end{equation}
	whose minimizer is the marginal field $v^*(x_t,c,t)=\mathbb{E}[u_t(x_t|x_0)\,|\,x_t,c]$. The significance of \eqref{eq:fm_loss} is that generative modeling reduces to supervised regression under a plain MSE loss, and the series of forward-process RL emerges as an attempt to keep that property while injecting a reward.
	
	Combining $v(x_t,c,t)=\frac{\hat{\epsilon}-x_t}{1-t}$ with Tweedie's formula $\nabla_{x_t}\log p_t(x_t|c) = -\hat{\epsilon}/t$~\citep{Efron01122011,song2021scorebased,pmlr-v202-song23a} expresses the field of \emph{any} endpoint law through the score of its noised marginal, $v=-\big(x_t + t\nabla_{x_t}\log p_t\big)/(1-t)$. Two laws noised by the same kernel would have fields differing only through the score of their ratio:
	\begin{equation}
		v^{*}(x_t,c,t) \;=\; v^{\old}(x_t,c,t) \;-\; \frac{t}{1-t}\,\nabla_{x_t}\log\frac{p^{*}_t(x_t\mid c)}{p^{\old}_t(x_t\mid c)},
		\label{eq:value_gradient_velocity}
	\end{equation}
	in which a reweighting of endpoints becomes an additive correction to the sampling field. It also disposes of the normalizer. $Z(c)$ in \eqref{eq:exp_tilt_sol} does not depend on $x_t$ and so leaves no trace in the gradient. The \emph{exact} target therefore never requires a partition function; \S\ref{subsec:objective_recovery} shows that the estimator actually implemented reintroduces one, and quantifies what that costs.
	
	This is still far from a trainable target: the ratio is taken between \emph{noised} marginals whereas the tilt is defined on endpoints, so the gradient is a posterior expectation rather than a per-sample quantity. Converting it is the remaining step, and since it is identical for every tilt we perform it once, in \S\ref{subsec:objective_recovery}.
	
	\section{From a Heuristic Linear Tilt to the Exact Simplex Projection}
	\label{app:diffusionnft_kl_equivalence}
	DiffusionNFT contrasts two implicit branches of the same model rather than forming a density ratio, so its update needs no likelihood ELBO~\citep{NEURIPS2025_3a10c465,Wallace_2024_CVPR}. For the same reason, it never states which divergence it controls. This section recovers that information and places the family inside \S\ref{subsec:method_prelims}.
	We find that a single DiffusionNFT update is a least-square regression onto a target of the advantage-weighted form, and that the reward enters that target \emph{linearly} rather than \emph{exponentially}. \S\ref{sec:nft_linear_tilt} puts the update in regression form and records the general statement that any endpoint reweighting induces such a target; \S\ref{subsec:objective_recovery} derives the reverse-KL counterpart, the exponential tilt, which coincides with the concurrently proposed FlowAWR objective; \S\ref{sec:chi2} identifies that a linear tilt is exactly what the Pearson $\chi^2$ penalty produces, but that its constrained optimum is a sparsemax projection of which the dense linear form is only the interior branch.
	
	\subsection{A Regression Form for the DiffusionNFT Update}
	\label{sec:nft_linear_tilt}
	
	Given an optimality probability $r\in[0,1]$, DiffusionNFT treats the reward-conditioned distribution as a mixture of a ``positive'' and a ``negative'' branch, and parameterizes either branch implicitly through the velocity difference it induces relative to $p^\old_t$. Because the two branches share the same noising family, that difference reduces to a difference of scores, which is the step we exploit below.%
	
	Concretely, the two branches are coupled through a single coefficient $\beta$. Factoring that coupling out leaves a sum of two ordinary regression terms (temporal and conditional arguments suppressed),
	\begin{equation}
		\mathcal{L}_{\text{NFT}}(\theta) = \beta^2 \mathbb{E} \bigg[ r \left\| v_\theta - \left(v^\old + \frac{1}{\beta}(u_t - v^\old)\right) \right\|^2 
		+ (1 - r) \left\| v_\theta - \left(v^\old - \frac{1}{\beta}(u_t - v^\old)\right) \right\|^2 \bigg].
		\label{eq:nft_factorized}
	\end{equation}
	Each term displaces the regression target from $v^\old$ along the residual $u_t-v^\old$, in opposite directions according to whether the sample is rewarded, and the population optimum of \eqref{eq:nft_factorized} is
	\begin{equation}
		v^*(x_t, c, t) = v^{\text{old}}(x_t, c, t) + \mathbb{E}_{x_0 \sim p^\old(\cdot | x_t, c)}  \frac{2r-1}{\beta} \left( u_t(x_t | x_0) - v^{\text{old}}(x_t, c, t) \right)  .
		\label{eq:nft_velocity}
	\end{equation}
	
	It already recovers the shape we desire: a sampling field plus a rewarded residual $u_t-v^\old$, that is not specific to DiffusionNFT, as \eqref{eq:value_gradient_velocity} forces it as soon as the target is any reweighting $p^*(x_0|c) \;=\; p^\old(x_0|c)\,w(c,x_0)$ of the sampling law, w.l.o.g., $w \ge 0$ and $\mathbb{E}_{x_0\sim p^\old(\cdot|c)}[w] = 1$. With $z(x_t,c,t)=\mathbb{E}_{x_0\sim p^\old(\cdot|x_t,c)}[w]$ being the posterior mean of the weight, substituting into \eqref{eq:value_gradient_velocity} and trading the posterior score for the residual collapses the correction into an expectation,
	\begin{equation}
		v^*(x_t, c, t) = v^\old(x_t, c, t) + \mathbb{E}_{x_0 \sim p^\old(\cdot | x_t, c)} \bigg[ \frac{w(c,x_0)}{z(x_t, c, t)} \,\big( u_t(x_t|x_0) - v^\old(x_t, c, t) \big) \bigg] .
		\label{eq:theorem_final}
	\end{equation}
	Both steps are elementary and \cref{app:rectification} gives them in full. In \eqref{eq:theorem_final}, the field is corrected by the residual reweighted by the \emph{relative} tilt $\kappa = w/z$, the weight a sample carries measured against the average weight of the endpoints compatible with the same $x_t$. Only that ratio appears, so any constant factor in $w$, a partition function in particular, is canceled out.
	Furthermore, since $v^\old$ is by construction the posterior mean of $u_t$, the residual has zero conditional mean, $\mathbb{E}_{p^\old(\cdot|x_t,c)}[u_t-v^\old]=0$, and therefore subtracting any constant $b$
	\begin{equation}
		\mathcal{A}(x_0,x_t) \;=\; \kappa\;-\; b
		\label{eq:advantage_baseline}
	\end{equation}
	leaves \eqref{eq:theorem_final} intact, so no value of $b$ is forced. Since $\mathbb{E}[w\mid x_t]=z$ by definition, the relative tilt has unit posterior mean and $b=1$ is the value that centers $\mathcal{A}$, turning it into a signed quality score in which positive advantages pull the field towards a sample and negative ones push it away. That is a convention instead of a derivation, and \S\ref{sec:ablation} uses the remaining freedom to choose $b$ so as to minimize variance. We quote the baseline $b$ below, explicitly with $b=1$ unless stated otherwise.
	
	Comparing \eqref{eq:nft_velocity} to \eqref{eq:theorem_final} now locates DiffusionNFT precisely. Its optimality probability $r=\tfrac12+z/(2C)$ is an affine mapping of the reward, where $z=(R-\bar R)/\sigma$ is the reward standardized within the prompt group and $C$ is a fixed constant. Substituting into \eqref{eq:nft_velocity} gives rise to
	\begin{equation}
		\mathcal{A}_{\text{NFT}} \;=\; \frac{2r-1}{\beta} \;=\; \frac{1}{\beta C}\,\frac{R(c,x_0)-\bar R(c)}{\sigma(c)} ,
		\label{eq:nft_weight_linear}
	\end{equation}
	so the weight DiffusionNFT applies is \emph{linear} in the reward, with no exponential anywhere, and $\beta C$ plays the role of a temperature. \Cref{app:nft_linear_derivation} corroborates that \eqref{eq:theorem_final} restores the same linear weight when the branch is established through Bayes' rule and Tweedie's formula.
	
	Nothing in that construction, however, claims the weight \emph{should} be linear, nor what the temperature should be, both are chosen heuristically. Set against the exponential tilt \eqref{eq:exp_tilt_sol} that reverse KL produces, DiffusionNFT is thus not solving the KL-regularized problem, which raises two questions: which penalty a linear tilt is its exact solution to, and whether the dense form from DiffusionNFT is that solution? \S\ref{sec:chi2} answers both, and the answer to the second is negative.
	
	\subsection{Returning to the KL Constraint: the Exponential Tilt}
	\label{subsec:objective_recovery}
	
	The machinery is indifferent to which tilt it is, so the reverse-KL case of \S\ref{subsec:method_prelims} costs nothing else. Its solution \eqref{eq:exp_tilt_sol} is the tilt $w\propto\exp(R/\gamma)$, and \eqref{eq:theorem_final} converts it into the reverse-KL instance
	\begin{equation}
		\kappa \;=\; \frac{\exp(R(c,x_0)/\gamma)}{\mathbb{E}_{x_0\sim p^\old(\cdot|x_t,c)}[\exp(R(c,x_0)/\gamma)]} ,
		\qquad \mathcal{A} = \kappa - b ,
		\label{eq:advantage_base}
	\end{equation}
	to be contrasted in \S\ref{sec:chi2} with its $\chi^2$ counterpart. The normalizer $Z(c)$ of \eqref{eq:exp_tilt_sol} has already been canceled in the ratio $\kappa$, and $\mathbb{E}_{p^\old(\cdot|x_t,c)}[\kappa]=1$ as always, so $b=1$ centers this advantage too.
	
	\Cref{eq:advantage_base} is still a population statement, as the reweighting $\kappa=w/z$ divides by the posterior mean $z(x_t,c,t)=\mathbb{E}_{x_0\sim p^\old(\cdot|x_t,c)}[w]$, an average over every endpoint that could have produced the same noised state. The linear tilt is in the same position and for the same reason: its threshold $\tau$ is fixed by \eqref{eq:waterfill}, itself being a posterior expectation. Both therefore call for the group-relative style estimator~\citep{2024arXiv240203300S,2024arXiv240214740A,2025arXiv250103262H}: draw a group of $G$ samples $\{x_0^{(i)}\}_{i=1}^G$ for the prompt $c$ and average over the grouped samples,
	\begin{equation}
		\kappa_i \;=\; \frac{w_i}{z_i} \approx \frac{\exp\left(R(c, x_0^{(i)})/\gamma\right)}{\frac{1}{G} \sum_{j=1}^G \exp\left(R(c, x_0^{(j)})/\gamma\right)} ,
		\qquad
		\mathcal{A}(x_0^{(i)}) = \kappa_i - b .
		\label{eq:Softmax_Advantage}
	\end{equation}
	It is worth being clear about what has just been swapped, because the group is not a sample from the posterior that \eqref{eq:advantage_base} asks about. The $x_0^{(i)}$ are drawn $\mathrm{i.i.d.}$ from the endpoint law $p^\old(\cdot|c)$, each from its own initial noise, so their average is unbiased for the normalizer $Z(c) \;=\; \mathbb{E}_{x_0\sim p^\old(\cdot|c)}[w]$ of \eqref{eq:exp_tilt_sol}, and not for the posterior mean $z(x_t,c,t)$, which conditions on the noised state. These two coincide only where the posterior equals the prior. Estimating $z$ itself is self-defeating rather than merely harder: the only samples available are prior draws, and reweighting them by $q(x_t|x_0^{(i)})$ provides a proposal whose efficiency degrades exponentially in higher dimension. At the latent size used here the reweighting concentrates on the single endpoint that generated $x_t$, hence $\hat z\to w_i$, $\kappa\to1$ and $\mathcal{A}\to0$, then the honest estimator annihilates the update. Substituting $Z(c)$ keeps the objective non-trivial, and its effect is characterized in \cref{app:z_vs_Z}. In broad terms, the net effect is simply a scalar reweighting $r_t(x_t)$. Writing $y$ for the regression target that \eqref{eq:Softmax_Advantage} induces,
	\begin{equation}
		\mathbb{E}\big[y \,\big|\, x_t\big] - v^\old = r_t(x_t)\,\big(v^*-v^\old\big),
		\quad
		r_t(x_t) \;=\; \frac{z(x_t,c,t)}{Z(c)} \;=\; \frac{p^*_t(x_t|c)}{p^\old_t(x_t|c)} \;>\;0 .
		\label{eq:Z_rescaling}
	\end{equation}
	The rescaling $r_t$ is centered at one. Its variance is bounded by $\mathbb{E}[\mathcal{A}^2]$, and decays towards the noisy end of the schedule, so the implemented target is displaced from the anchor in exactly the direction of the optimal field and only its step length is affected. Refer to \cref{app:z_vs_Z} for a rigorous proof.%
	
	Up to now, one expectation is still alive, and it is the ordinary one that any flow-matching loss carries: the average over the noise level $t$ and the renoising draw $\epsilon$. Estimating it by a single Monte-Carlo pair per sampled endpoint turns \eqref{eq:theorem_final} into a per-sample least-square objective,
	\begin{equation}
		\label{eq:group_loss}
		\begin{split}
			&\mathcal{L}(\theta)=\frac{1}{G}\sum_{i=1}^{G}\mathbb{E}_{t,\epsilon}\bigg[  \Big\| v_\theta(x_t^{(i)},t)-\Big(v^\old(x_t^{(i)},t)+\mathcal{A}(x_0^{(i)})\cdot\big(u_t(x_t^{(i)}|x_0^{(i)})-v^\old(x_t^{(i)},t)\big)\Big)\Big\|^2 \bigg],
		\end{split}
	\end{equation}
	in which every term is computable from the pair at hand. This substitution differs in kind from the previous one. It replaces an exact target by a \emph{stochastic} one, and unlike the passage from $z$ to $Z$ it is free. Conditionally on $x_t$ the drawn target has the same mean as the quantity it stands for, so the objective differs by a constant independent of $\theta$. The gradients and minimizer are unchanged, and only the variance of the gradient estimator is affected. \Cref{app:minimizer} affords the argument, which is the same tower-property observation that licenses conditional flow matching itself. The constant is identified as the conditional variance of the target, which is also what the baseline of \eqref{eq:optimal_cv} tends to reduce later. Combining the two substitutions, \eqref{eq:group_loss} is an unbiased stochastic objective for the rescaled field $v^\old+r_t(v^*-v^\old)$ of \eqref{eq:Z_rescaling}, rather than for $v^*$ itself.
	
	\Cref{eq:group_loss} with the softmax weight in \eqref{eq:Softmax_Advantage} echoes the objective proposed concurrently by FlowAWR~\citep{2026arXiv260630376F}, which we thus recover as the reverse-KL corner of~\eqref{eq:density_ratio} rather than as a separate algorithm. The Pearson $\chi^2$ corner, and the sense in which it is the one DiffusionNFT occupies, is the subject of \S\ref{sec:chi2}.
	
	\subsection{The Exact Solution: Pearson \texorpdfstring{$\chi^2$}{chi-squared} and the Sparsemax Tilt}%
	\label{sec:chi2}
	
	The linear reward weighting that \S\ref{sec:nft_linear_tilt} extracted from the DiffusionNFT update is not an artefact of that particular construction, it is what the problem of \S\ref{subsec:method_prelims} returns for a different generator. Retaining the family of \eqref{eq:density_ratio} and replacing $f_{\mathrm{KL}}(\rho)=\rho\log\rho$ by the Pearson $\chi^2$ generator $f_{\chi^2}(\rho) = \tfrac12(\rho-1)^2$, so that $D_{\chi^2}$ is the half second moment of $\rho-1$ under $\pi^\old$, the objective becomes
	\begin{equation}
		\max_{\pi_\theta} \mathbb{E}_{s \sim \mathcal{D}} \left[ \mathbb{E}_{a \sim \pi_\theta(\cdot|s)} R(s, a) - \gamma D_{\chi^2}(\pi_\theta(\cdot|s) || \pi^\old(\cdot|s)) \right].
		\label{eq:chi2_constrained_objective}
	\end{equation}
	\begin{figure}[t]
		\vspace{-3.5em}
		\begin{center}
			\includegraphics[width=\linewidth]{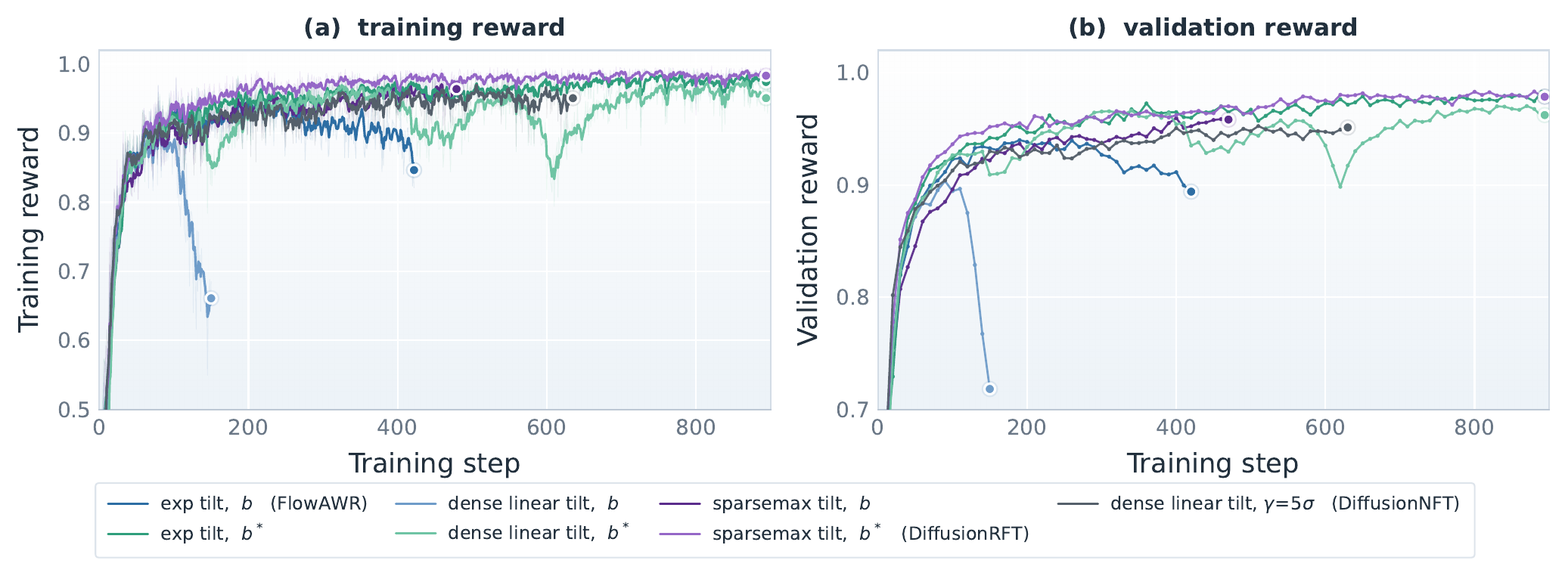}
		\end{center}
		\vspace{-1.5em}
		\caption{\textbf{Tilt $\times$ baseline grid.} Three tilts (exponential, dense linear, sparsemax) crossed with two baselines (canonical constant $b$, variance-minimizing control variate $b^*$). The sparsemax tilt reduces to the dense linear tilt when the truncation is inactive, i.e., $\min_i\mathcal{A}>-1$. Yet, this criterion fails on $86.8\%$ of the dense run's steps. The optimal control variate $b^*$ stabilizes and strengthens every tilt.}
		\label{fig:tilt-baseline}
		\vspace{-.5em}
	\end{figure}Unlike the reverse KL, whose solution is less constrained in practice, the $\chi^2$ problem must be solved over $\mathcal{K} = \{\rho: \mathbb{E}_{\pi^\old}[\rho] = 1, \rho \ge 0\}$, which imposes normalization \emph{and} non-negativity. The objective in \eqref{eq:chi2_constrained_objective} is strictly concave in $\rho$, and $\mathcal{K}$ convex and non-empty, so the maximizer exists, is unique, and is fully characterized by the Karush--Kuhn--Tucker conditions~\citep{boyd2004convex}. Complementary slackness on the non-negativity multiplier splits it into an interior branch and a boundary branch, and coalescing them as in \cref{app:chi2_derivation} yields the optimal terminal distribution
	\begin{equation}
		p^*(x_0 | c) \;=\; p^\old(x_0 | c)\,\cdot\,\frac{1}{\gamma}\Big[\, R(c, x_0) - \tau(c) \,\Big]_+ ,
		\qquad [\,z\,]_+ = \max(z,0),
		\label{eq:lin_tilt_sol}
	\end{equation}
	in which the threshold $\tau(c)$ is not free but is pinned by the normalization constraint,
	\begin{equation}
		\Phi(\tau) \;\equiv\; \mathbb{E}_{x_0\sim p^\old(\cdot|c)}\big[\,(R(c,x_0)-\tau)_+\,\big] \;=\; \gamma .
		\label{eq:waterfill}
	\end{equation}
	Since $\Phi$ is convex, continuous and strictly decreasing on $(-\infty, R_{\max})$ with $\Phi(R_{\max})=0$, \eqref{eq:waterfill} has a unique root, therefore~\eqref{eq:lin_tilt_sol} admits a transparent reading: $\tau$ stands for a \emph{water level} determined by the regularization budget $\gamma$, samples above it receive probability mass in proportion to their excess reward, while samples below it receive none. Note that no partition function survives, since \eqref{eq:waterfill} already enforces $\mathbb{E}_{p^\old}[\rho]=1$: where the exponential tilt has to estimate $Z(c)$ by a group average, the projection obtains it from $\tau$ at no cost. This does not exempt it from the posterior normalizer $z$ of \eqref{eq:theorem_final}, that each member of the family bypasses rather than calculates.
	
	Two circumstances are separated as follows, contingent upon the KKT condition. When $\gamma$ is large relative to the spread of the reward, every sample lies above the water level, the constraint $\rho\ge0$ is relaxed, and \eqref{eq:waterfill} reduces to $\Phi(\tau)=\bar R-\tau$ so that $\tau = \bar R - \gamma$. Substituting back recovers the familiar centered form
	\begin{equation}
		p^*(x_0|c) \;=\; p^\old(x_0|c)\left(1 + \frac{R(c,x_0)-\bar R(c)}{\gamma}\right),
		\qquad
		\mathcal{A}_{\chi^2}(x_0) \;=\; \frac{R(c,x_0)-\bar R(c)}{\gamma},
		\label{eq:lin_tilt_interior}
	\end{equation}
	which is exactly the linear weighting of \S\ref{sec:nft_linear_tilt}, and also the group-centered advantage used by GRPO-style estimators. This regime is characterized sharply by
	\begin{equation}
		\gamma \;>\; \bar R(c) - \operatorname*{ess\,inf} R(c,\cdot)
		\quad\Longleftrightarrow\quad
		\kappa \;>\; 0
		\;\;(\text{equivalently } \mathcal{A}_{\chi^2} > -b)
		\;\;\text{ for every sample},
		\label{eq:validity}
	\end{equation}
	whose right-hand form costs nothing to monitor during training and will be leveraged in \S\ref{sec:design_space} to decide empirically which regime a run is in. In the second regime the water level rises above the weakest samples, $\rho$ saturates at zero for them, and \eqref{eq:lin_tilt_interior} is no longer the solution of \eqref{eq:chi2_constrained_objective}.

	Criterion \eqref{eq:validity} settles one more thing: why DiffusionNFT's heuristic is significant. Its weight \eqref{eq:nft_weight_linear} is \eqref{eq:lin_tilt_interior} at temperature $\beta C\sigma$, and a group-standardized reward is bounded by $\sqrt{G-1}$ whatever the rewards are, so \eqref{eq:validity} holds for the whole family exactly when $G\le 1+(\beta C)^2$ (see \cref{app:chi2_derivation} for derivation). DiffusionNFT satisfies this bound only with its designated hyperparameters, while the sparsemax projection is feasible at every temperature and every $G$.

	That the two divergences differ here is structural. The optimality condition reads $\rho = (f')^{-1}\!\big((R-\lambda)/\gamma\big)$, so $f'$ alone decides the shape of the tilt: $f'_{\text{KL}}(\rho)=\log\rho+1$ diverges at $\rho\to0^+$ and acts as a logarithmic barrier, so $\rho\ge0$ is never active, and inverting $f'$ produces the exponential tilt; whereas $f'_{\chi^2}(\rho)=\rho-1$ is finite at the boundary, which is reachable, and inverting $f'$ produces a linear tilt with the constraint active when $(R-\lambda)/\gamma < -1$. Every non-barrier $f$-divergence inherits this behavior, so the truncation in \eqref{eq:lin_tilt_sol} is generic to the linear family rather than peculiar to $\chi^2$. Feeding this tilt through \eqref{eq:theorem_final} gives the optimal velocity field with $\kappa=[R-\tau]_+/\gamma$; \cref{app:chi2_derivation} carries out the KKT solution, the intermediate marginals and the projection step in full.
	
	Applying to \eqref{eq:lin_tilt_sol} the same group estimator in \S\ref{subsec:objective_recovery} makes its structure explicit. Let $w_i = \frac{\rho_i}{G}$ be the probability that $\pi_\theta$ assigns to $x_0^{(i)}$, so that $\mathcal{K}$ becomes the probability simplex $\Delta^{G-1}$. The empirical form of \eqref{eq:chi2_constrained_objective} is then $\max_{w\in\Delta}\,\langle w, R\rangle - \tfrac{\gamma G}{2}\|w - \tfrac1G\mathbf{1}\|^2$, and completing the square identifies it as a Euclidean projection onto the simplex:
	\begin{equation}
		w^\star=\operatorname{sparsemax}\!\Big(\tfrac1G\mathbf{1}+\tfrac{R}{\gamma G}\Big),
		\qquad
		\mathcal{A}_{\chi^2}(x_0^{(i)})=\kappa - b = G\,w^\star_i - b =\frac{\big[R(c,x_0^{(i)})-\tau\big]_+}{\gamma}-b ,
		\label{eq:sparsemax_adv}
	\end{equation}
	where $\tau$ now solves $\sum_{j}(R(c,x_0^{(j)})-\tau)_+=\gamma G$, computable in $\bigO(G\log G)$ by sorting the group rewards. The two tilts are thus in exact correspondence: the exponential tilt is $\operatorname{softmax}$ on the simplex and the linear tilt is the $\operatorname{sparsemax}$ of \citet{pmlr-v48-martins16}, the entropic and Euclidean projections of the same linear objective, and \eqref{eq:lin_tilt_interior} is the sparsemax solution restricted to its dense regime. This dense form is also adopted by concurrent works on flow RL, either directly~\citep{2026arXiv260526013K} or after reshaping the advantage~\citep{2026arXiv260510937S}. \Eqref{eq:sparsemax_adv} reveals what such choices approximate, and \eqref{eq:validity} unveils when the approximation is exact.

	\paragraph{Remark.} The three algorithms are distinct by the choice of $w$ and by nothing else: the reverse KL yields the softmax weight \eqref{eq:Softmax_Advantage}, the Pearson $\chi^2$ the sparsemax weight \eqref{eq:sparsemax_adv}, and the linear tilt that DiffusionNFT adopts heuristically is the interior branch in \eqref{eq:lin_tilt_interior} of the latter. What the derivation does \emph{not} fix is everything outside that choice, and \S\ref{sec:design_space} takes it up.
	
	\begin{figure}[t]
		\vspace{-3.5em}
		\begin{center}
			\includegraphics[width=\linewidth]{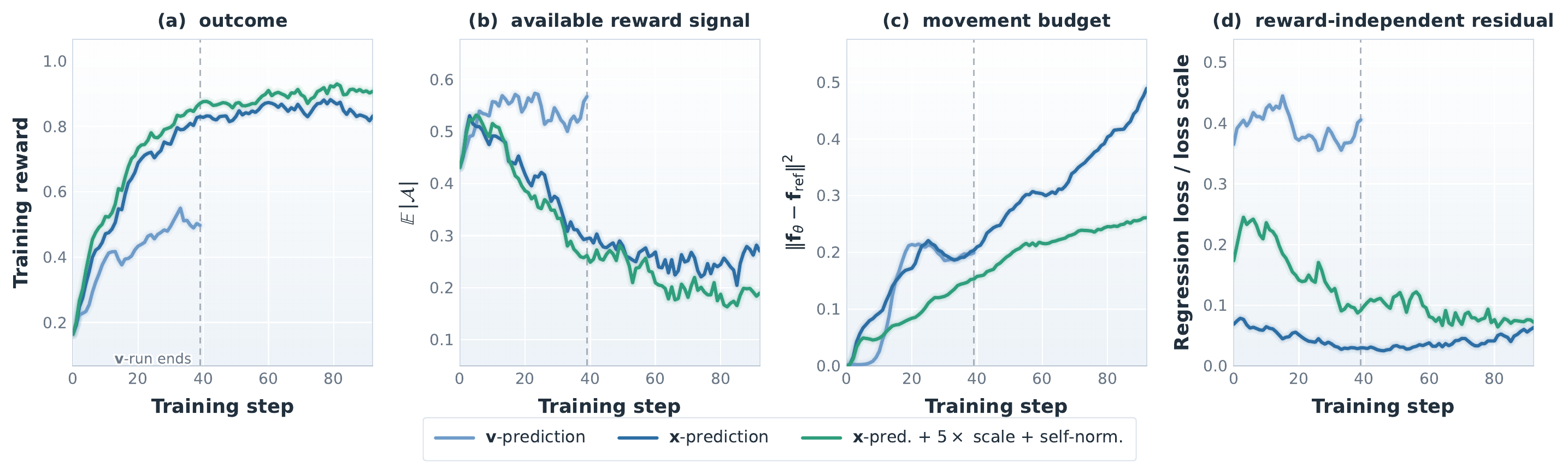}
		\end{center}
		\vspace{-1.5em}
		\caption{\textbf{Loss geometry: the $\vv$-space update is misdirected, not too small.} More advantage signal (b) and largest movement budget (c), but an order-of-magnitude larger residual (d), and correspondingly less reward (a). The dashed line marks the matched horizon among these runs.}%
	\label{fig:loss-dynamics}
	\vspace{-.5em}
	\end{figure}
	
	\section{Design Space Navigation}
	\label{sec:design_space}
	
	The unified optimization objective is $\mathcal{L}(\vtheta) = \mathbb{E}_{t,\epsilon} \Vert \vf_\vtheta(\vx_t) - \vf_{\text{target}} \Vert^2$, where
	\begin{equation}
		\vf_{\text{target}} = \vf_{\text{anchor}} + \mathcal{A} (\vf_{\text{gt}} - \vf_{\text{old}}), \qquad \vx_t = (1-t)\,\tilde{\vx}_0 + t\,\epsilon.
		\label{eq:unified_target}
	\end{equation}
	
	Here $\tilde{\vx}_0$ is the clean sample returned by the sampler through the whole ODE denoising trajectory, and $\epsilon$ is the draw used to renoise that sample back to timestep $t$. $\vf_{\text{old}}$ is the rollout policy responsible for sampling the trajectory, $\vf_{\text{anchor}}$ is the field the target is displaced from (the value the regression falls back to when the advantage vanishes, set to $\vf_{\text{old}}$ by \eqref{eq:theorem_final} but not forced to be), and $\vf_{\text{gt}}$ is the ground truth, which could be $\vx$-, $\epsilon$-, or $\vv$-prediction~\citep{Li_2026_CVPR,NEURIPS2020_4c5bcfec,lipman2023flow}. There exists a well-documented source of discrepancy in diffusion training, where the choice of parameterization and the associated loss weighting substantially determines convergence speed and stability~\citep{NEURIPS2022_a98846e9,NEURIPS2023_ce79fbf9,pmlr-v235-esser24a}.%
	
	Reading \eqref{eq:unified_target} together with \eqref{eq:group_loss} exposes four design choices that the derivation of \S\ref{app:diffusionnft_kl_equivalence} leaves open and \S\ref{sec:ablation} varies one at a time:
	\begin{enumerate}[leftmargin=2em,itemsep=0.15em,topsep=0.3em]
		\item[\textbf{(i)}] \textbf{loss function}: the variable space $\vf$ in which the regression is materialized, and the weighting across noise levels that this choice prefers;
		\item[\textbf{(ii)}] \textbf{reward shaping}: which tilt produces the weight $\kappa$, and which baseline $b$ is subtracted from it;
		\item[\textbf{(iii)}] \textbf{anchor point}: whether $\vf_{\text{anchor}}$ is the rollout policy $\vf_{\text{old}}$, refreshed every epoch, or the pretrained one $\vf_{\text{ref}}$, held fixed for the whole run; they agree at the first update and diverge from then on;
		\item[\textbf{(iv)}] \textbf{experience buffer}: the law of the renoising timestep $t$ at which the loss is evaluated.
	\end{enumerate}
	
	\begin{figure}[t]
	\vspace{-3.5em}
	\begin{center}
		\includegraphics[width=\linewidth]{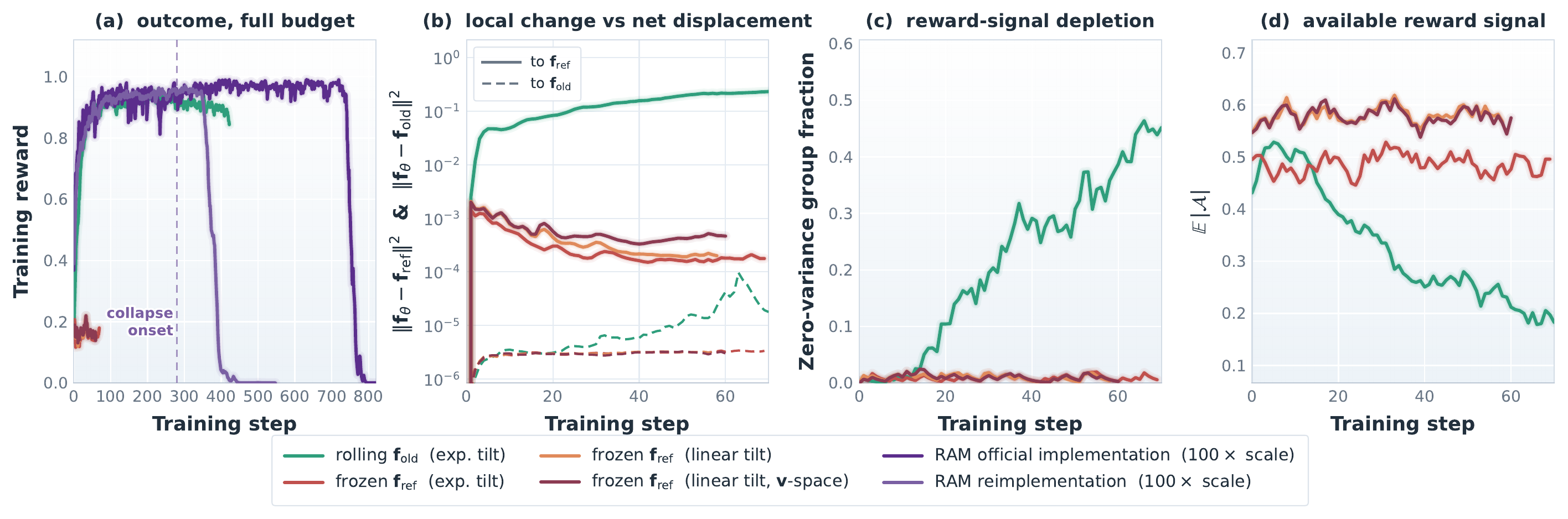}
	\end{center}
	\vspace{-1.5em}
	\caption{\textbf{Anchor-lock is a displacement failure, not a signal failure.} (a) spans the full budget, where both the official and reimplemented RAM escape the lock with the aid of a $100\times$ advantage scale but then collapse. (b)--(d) are restricted to the matched window in which the frozen-anchor runs coexist: dashed, non-zero step updates; solid, no effective accumulation (b) and no prompt group is ever solved (c), even though the advantage is \emph{larger} than under a rolling anchor (d). The three frozen-anchor curves overlap in every panel, which pinpoints the anchor as the root cause.}
	\label{fig:anchor-dynamics}
	\vspace{-.5em}
	\end{figure}
	
	\subsection{Experiment settings}
	\label{subsec:exp_settings}
	For a fair comparison and reproducibility, we follow DiffusionNFT's codebase to adapt other follow-up variants. The training and evaluation dataset is GenEval~\citep{NEURIPS2023_a3bf71c7}, that measures compositional spatial reasoning in text-to-image tasks. All experiments are conducted with the Stable Diffusion 3.5-Medium backbone~\citep{pmlr-v235-esser24a} at a $512\times512$ resolution. Other experimental details and additional results on Pickscore~\citep{NEURIPS2023_73aacd8b} are provided in \cref{app:impl_details}.
	
	\subsection{Ablation studies}
	\label{sec:ablation}
	
	\paragraph{loss function} We start from \eqref{eq:group_loss}, a widely acknowledged exponential tilt, to launch our exploration. We find that the $\vv$-prediction described in prior literature~\citep{zheng2026diffusionnft,2026arXiv260630376F} converges slowly in training, agreeing with the gap between theoretical justification and empirical implementation, as mentioned by~\citet{2026arXiv260526013K}. Consequently, we adopt $\vx$-prediction to expedite training hereafter, in light of a sharp gap displayed in \figref{fig:loss-dynamics}(a). Furthermore, certain subtle implementation details decide training dynamics. Particularly, the loss scaling factor and self-normalization accelerate training convergence. These points of interest are elaborated in another contemporary work in the context of diffusion distillation~\citep{2026arXiv260519256K}.%
	
	The training dynamics in \figref{fig:loss-dynamics} show that the gap concerns the update direction, not magnitude. At a matched horizon the $\vv$-prediction run carries the \emph{largest} advantage and spends the \emph{largest} movement budget, but gains the \emph{least} reward, while its regression loss is an order of magnitude higher than $\vx$-prediction. An empirical decomposition constrains several contributors, e.g. advantage scale, prediction units, and residual excess, only loosely, leaving room for the change in the irreducible floor alone to account for the observed loss ratio. In particular, the $\vx$-space loss eschews the coefficient $t^{-2}$ with which the $\vv$-space loss uplifts the endpoint's irreducible inflation. \Cref{app:ablation_details} delves deeper into the rationale. The resulting $+0.34$ gain in the terminal training reward is pronounced, so we fix this loss geometry throughout rather than treating it as one more knob.%
	
	\paragraph{reward shaping}
	
	According to the previous section, the exponential and linear tilt are both viable paths to diffusion RL outcomes. However, to minimize the variance of $(\mathcal{A}-b)\delta$ during RL training, where $\delta=u_t-v^\old$ for brevity, the analytic solution of this control variate is 
	\begin{equation}
		b^* = \mathbb{E} \left[\mathcal{A}\Vert \delta \Vert^2\right]\, /\, \mathbb{E} \Vert \delta \Vert^2.
		\label{eq:optimal_cv}
	\end{equation}
	Different from the pilot experiments above, we extend training iterations to over $10\times$.  Three conclusions survive the full $3\times2$ grid in \figref{fig:tilt-baseline}. \textbf{(1)} Collapse requires \emph{both} a large effective update and the absence of a control variate, since either the projection or $b^*$ prevents it alone. Once $b^*$ is in place the three tilts are all stabilized. \textbf{(2)} These two strategies are complementary rather than interchangeable: the projection keeps the linear tilt inside the region where it is the optimum, while $b^*$ advances the envelope of any tilt. Their combination, DiffusionRFT, is the safest and strongest configuration, at $0.983$ evaluation reward. \textbf{(3)} The dashed reference line of DiffusionNFT also sits at the dense linear tilt cell, at a hard-coded temperature $\gamma=5\sigma$, against $\gamma=\sigma$ for the plain dense linear tilt. The gap between the two curves is predicated on \eqref{eq:validity}: at $\gamma=5\sigma$ and $G=24$ the interior branch is always legal, while at $\gamma=\sigma$ it is violated on $86.8\%$ of steps. \Cref{app:ablation_details} affords a deeper account of the validity diagnostics.
	
	\paragraph{anchor point}
	
	DiffusionNFT anchors on a rolling copy, $\vf_{\text{anchor}}=\vf_\old\leftarrow(1-\mu_i)\vf_\old+\mu_i\vf_\theta$, trading staleness for stability. RAM instead anchors on the frozen pretrained model $\vf_{\text{ref}}$. Replacing $\vf_\old$ by $\vf_{\text{ref}}$ alone produces an \emph{anchor-lock} phenomenon: the training reward of three frozen-anchor variants (with different tilt and regression space) rises to below $0.25$ and stagnates, so the ceiling is a property of the anchor and not of one particular setup.
	
	The culprit of lock is limited displacement $\|\vf_\theta-\vf_{\text{anchor}}\|^2$, not a starved signal. In \figref{fig:anchor-dynamics}, the frozen-anchor runs carry \emph{larger} advantages than the rolling-anchor one, yet they never leave the pretrained field, even though the local change is non-negligible from the outset. With the anchor fixed the target is $\vf_{\text{ref}}+\mathcal{A}\,\delta$, so each update re-centers the policy to $\vf_{\text{ref}}$ and the attainable displacement is bounded by the advantage-weighted residual itself, whereas a rolling anchor $\vf_\old$ tracks $\vf_\theta$ closely.
	Escaping the lock requires a larger update, and both RAM implementations, official and our reproduction of its skeleton, show that the escape is temporary. They peak at $0.977$ and $0.961$ and then fall to zero. The frozen anchor therefore has no safe operating point under our test: small updates lock, large ones diverge. In contrast, a rolling anchor needs no such handling, because re-centering keeps the displacement of order one at every epoch.
	\Cref{app:ablation_details} analyzes the mechanism in full.
	
	\begin{figure}[t]
		\vspace{-3.5em}
		\begin{center}
			\includegraphics[width=\linewidth]{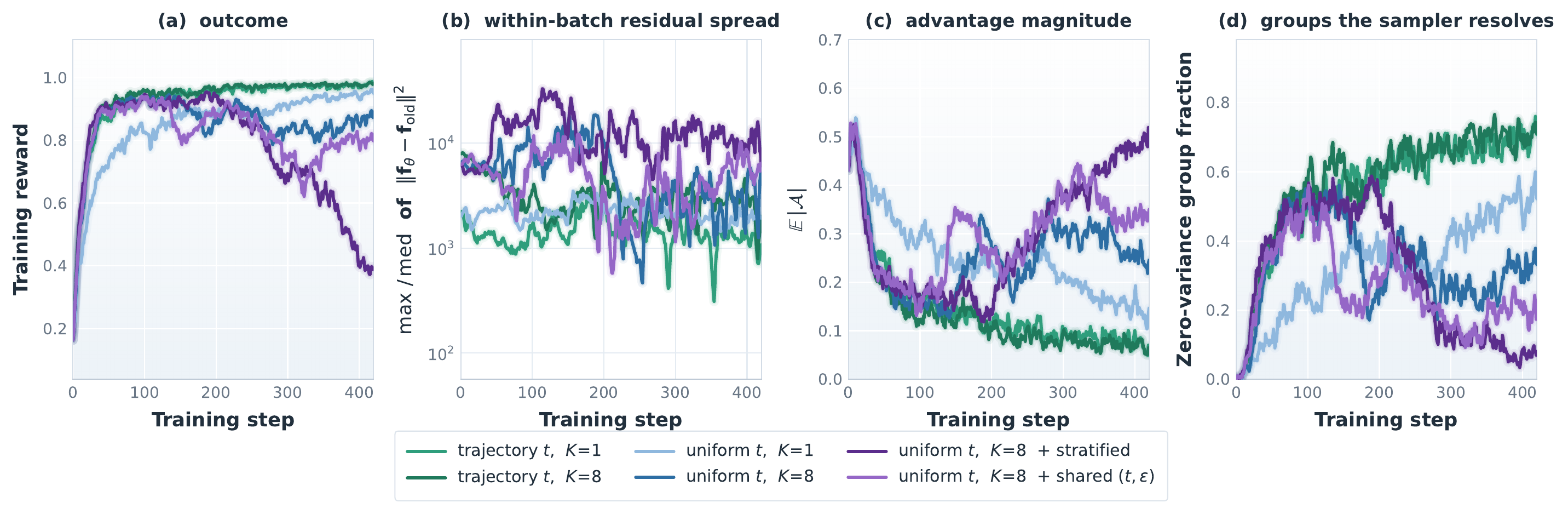}
		\end{center}
		\vspace{-1.5em}
		\caption{\textbf{Decoupling the renoising timestep makes the batch heavy-tailed, and stratification makes it worse.} (a) Training reward. (b) Within-batch spread of the policy-EMA gap $\|\vf_\theta-\vf_\old\|^2$ on a logarithmic axis: lowest for the trajectory-coupled schedule, $\sim2\times$ once the timestep is decoupled, and $\sim8\times$ once the draws are stratified. (c) Advantage magnitude: the trajectory-coupled run declines as it solves its prompt groups, while the decoupled runs remain high. (d) Fraction of prompt groups whose rewards have zero variance, which carry no advantage. A uniform distribution spends its evaluations on uninformative states the sampler hardly visits instead of the ones it resolves.}
		\label{fig:buffer-dynamics}
		\vspace{-.5em}
	\end{figure}
	
	\paragraph{experience buffer}
	
	The training states $\vx_t$ are renoised from the generated clean samples, where the renoising timesteps are tied to the backward denoising trajectory, which is reminiscent of the classic replay buffer in RL~\citep{mnih2015human}. Differently, RAM decouples the forward renoising timestep schedule from the denoising trajectory: renoise timesteps are drawn from a uniform distribution and repeated $K=8$ times, accumulating an $8\times$ experience buffer for one-step optimization.
	
	RAM's sampling conflates two factors at once, the law of $t$ and the number of copies $K$. Crossing the two achieves evaluation rewards of $0.972$ \& $0.975$ for the trajectory-coupled law at $K=1$ \& $K=8$, against $0.964$ \& $0.953$ for the uniform law, the last of these losing a further $0.086$ before the budget ends (\figref{fig:buffer-dynamics}). Expanding sampling $K$ is therefore neutral and it is the law of $t$ that counts. The underlying cause is that decoupling alters the \emph{support} of $t$: the sampler-oriented distribution never visits $t<0.43$ (\figref{fig:traj-t}), while $t\sim\mathcal{U}[0,1]$ places $43\%$ of the loss evaluations there. It also clarifies why two variance-reduction remedies, stratified sampling and shared $(t, \epsilon)$ sampling across copies~\citep{2026arXiv260423380T}, still fail, and why stratification, that enforces the uniform coverage, is even the more damaging of these two. Renoising along the generating trajectory only once ($K=1$) keeps the training states on the support the policy occupies, which is both cheaper and better conditioned, so we retain it by default. \Cref{app:ablation_details} incorporates more in-depth diagnostics.
	
	\section{Conclusion}
	
	Regression-based reinforcement learning for diffusion models admits a unified formulation with a free parameter. The convex generator in the divergence penalty determines the particular choice of this parameter: an exponential tilt for FlowAWR; a linear tilt whose exact solution, the sparsemax projection, leads to our proposed DiffusionRFT; DiffusionNFT heuristically adopts the dense linear form that lies in the interior branch of the latter. The empirical picture favors getting the tilt done right. Standing on its own, the projected linear tilt performs the best among them, and it is the only candidate that does not surrender its peak reward; a variance-minimizing baseline then stabilizes and improves all three tilts, and the two together approach the fastest convergence and highest reward. Besides the tilt choice, what bears emphasis includes $\vx$-prediction with appropriate loss calibration, an online cumulative anchor point, and a backward-aligned forward renoising schedule.%
	
	\label{sec:main_end}

	\subsection*{AI use statement}
	In this work, we used generative AI tools for manuscript polish and \LaTeX{} checks during revision.
	We have not used generative AI tools for anything else. We have reviewed all AI-assisted work.  We take responsibility for the final content of this work,
	including text, claims or artifacts produced with the aid of generative AI.
	
	\subsection*{Reproducibility statement}
	
	We believe the experimental details in \cref{app:impl_details} are ready for reproduction. In addition, we would release the code upon publication to foster future research.

	\bibliography{iclr2027_conference}
	\bibliographystyle{iclr2027_conference}
	
	\appendix
	\crefalias{section}{appendix}
	
	\section{Related work}
	\label{app:related}
	\subsection{Diffusion Generative Modeling}
	
	Inspired by thermodynamics~\citep{pmlr-v37-sohl-dickstein15} in physics, the advent of diffusion models occurred one decade ago. In the following years, after briefly overshadowed by Generative Adversarial Networks (GANs)~\citep{2014arXiv1406.2661G}, improved empirical practice~\citep{NEURIPS2020_4c5bcfec} and a firmer theoretical footing~\citep{song2021scorebased} ignited the resurgence of diffusion generative modeling. The system is further scaled up with advanced architecture engineering~\citep{Rombach_2022_CVPR,Peebles_2023_ICCV}, distribution coupling~\citep{lipman2023flow,liu2023flow}, network capacity reallocation~\citep{NEURIPS2022_a98846e9,pmlr-v235-esser24a,NEURIPS2023_ce79fbf9}, and higher-order solvers~\citep{song2021denoising,NEURIPS2022_260a14ac}. Embracing diffusion as the generative modeling backend, recent years have witnessed an unprecedented achievement across a variety of visual generative applications, such as text-to-image~\citep{2025arXiv250802324W,2026arXiv260510730Z}, video~\citep{2025arXiv250609113G,2026arXiv260414148T}, and 3D geometry~\citep{2024arXiv241102293Y,2025arXiv250112202Z}. Thanks to the pixel-level generative modeling, diffusion models permit a supervised learning MSE loss to be established and optimized stably (much more than GANs' brittle minimax game). Nevertheless, a coin always has two sides: this training objective also falls short of straightforwardly modeling the sample-level human preference, such as visual aesthetics, physical plausibility, among others.
	
	\subsection{Diffusion Reinforcement Learning}
	
	Reinforcement learning has gained tremendous traction in Large Language Models (LLMs), contributing to both alignment with human preference~\citep{NEURIPS2022_b1efde53} and incentivization of Chain-of-Thought (CoT) reasoning~\citep{2025Natur.645..633G}. The transfer of RL in autoregressive modeling to diffusion models poses a challenge, as there is no well-defined tractable likelihood formulation in diffusion modeling. One paradigm converts the deterministic ODE solver into a stochastic SDE, in order to frame each transition as a step of a Markovian Decision Process (MDP) and supplies the exploration noise that policy-gradient estimators require~\citep{NEURIPS2025_3a10c465,Wang_2026_CVPR}. Another strand sidesteps the likelihood calculation via directly back-propagating a differentiable reward through the sampler~\citep{NEURIPS2023_33646ef0,clark2024directly,2024arXiv240708737P}. Instead, DiffusionNFT innovatively presents the first diffusion-native RL prototype, through the lens of implicit contrastive learning between positive and negative representation fields~\citep{zheng2026diffusionnft}. However, the underlying insight of its reward-integrated regression target is not disclosed. Likewise, several recent works~\citep{2026arXiv260630376F,xue2026advantage,2026arXiv260526013K} also lie in the lineage of advantage-weighted regression~\citep{2019arXiv191000177P}, where a reward-tilted distribution is fitted by weighted least squares rather than followed by a gradient. Mindful of its design elegance and the built-in compatibility with diffusion loss, we first link the DiffusionNFT formulation to the classic constrained reward maximization problem, and then venture as far as possible into the design space.

	\section{Derivation of the reweighted velocity field}
	\label{app:rectification}
	
	This appendix derives \eqref{eq:theorem_final}. Let $p^*(x_0|c)=p^\old(x_0|c)\,w(c,x_0)$ be any tilt, as in \S\ref{sec:nft_linear_tilt}. The derivation has four steps: push the tilt through the noising kernel, differentiate the resulting log-ratio, evaluate the conditional score in closed form, and collect the prefactors.
	
	\paragraph{Step 1: noised marginals.}
	Both laws are noised by the same kernel $p(x_t|x_0)$, so
	\begin{equation}
		p_t^*(x_t | c)=\int p(x_t|x_0)\,p^*(x_0|c)\,\mathrm{d}x_0
		=\int p(x_t|x_0)\,p^\old(x_0|c)\,w(c,x_0)\,\mathrm{d}x_0 .
		\label{eq:app_marginalize}
	\end{equation}
	Bayes' rule for the reference process reads $p(x_t|x_0)\,p^\old(x_0|c)=p^\old_t(x_t|c)\,p^\old(x_0|x_t,c)$. Substituting it into \eqref{eq:app_marginalize} and taking the $x_0$-independent factor out of the integral yields
	\begin{equation}
		\begin{split}
			p_t^*(x_t | c)
			&= p_t^\old(x_t | c)\int p^\old(x_0|x_t,c)\,w(c,x_0)\,\mathrm{d}x_0
			= p_t^\old(x_t | c)\cdot z(x_t, c, t), \\
			z \;&\equiv\; \mathbb{E}_{x_0\sim p^\old(\cdot|x_t,c)}\big[w(c,x_0)\big].
		\end{split}
		\label{eq:p_t^*}
	\end{equation}
	The tilt on endpoints thus becomes a tilt on noised marginals, but by the \emph{posterior mean} of $w$ rather than by $w$ itself. This is the only place the noising kernel is used, and it is why $z$ cannot be evaluated per sample.
	
	\paragraph{Step 2: differentiate the log-ratio.}
	By \eqref{eq:p_t^*} the ratio of noised marginals is exactly $z$, so \eqref{eq:value_gradient_velocity} becomes
	\begin{equation}
		v^{*}-v^{\old}
		= -\frac{t}{1-t}\,\nabla_{x_t}\log\frac{p^{*}_t(x_t|c)}{p^{\old}_t(x_t|c)}
		= -\frac{t}{1-t}\,\nabla_{x_t}\log z
		= -\frac{t}{1-t}\,\frac{\nabla_{x_t}z}{z} .
		\label{eq:app_grad_phi}
	\end{equation}
	For the numerator, differentiation of \eqref{eq:p_t^*} under the integral sign is legitimate whenever $w$ is $p^\old$-integrable and the Gaussian kernel dominates, which holds here since $w\ge0$ has unit mean. Multiply $w$ by the posterior and then differentiate the integral:
	\begin{equation}
		\begin{split}
			\nabla_{x_t}z
			&= \int \nabla_{x_t}p^\old(x_0|x_t,c)\;w\,\mathrm{d}x_0 \\
			&= \int p^\old(x_0|x_t,c)\,\nabla_{x_t}\!\log p^\old(x_0|x_t,c)\;w\,\mathrm{d}x_0 \\
			&= \mathbb{E}_{x_0\sim p^\old(\cdot|x_t,c)}\Big[w\,\nabla_{x_t}\!\log p^\old(x_0|x_t,c)\Big],
		\end{split}
		\label{eq:app_logderiv}
	\end{equation}
	in which the middle equality is the log-derivative identity $\nabla \pi=\pi\,\nabla\log \pi$. Dividing by $z$ converts \eqref{eq:app_logderiv} into an expectation of the conditional score weighted by the relative tilt $w/z$.
	
	\paragraph{Step 3: conditional score in a closed form.}
	Bayes' rule once more, now in the form 
	\begin{equation}
		\log p^\old(x_0|x_t,c)=\log p(x_t|x_0)+\log p^\old(x_0|c)-\log p^\old_t(x_t|c).
	\end{equation}
	We kill the middle term under $\nabla_{x_t}$ and leave the other two scores. The forward noising kernel of \S\ref{subsec:tilt_to_velocity} is $x_t|x_0\sim\mathcal{N}\big((1-t)x_0,\,t^2 I\big)$, whose score is affine,
	\begin{equation}
		\nabla_{x_t}\log p(x_t|x_0) = -\frac{x_t-(1-t)x_0}{t^{2}} = \frac{(1-t)x_0-x_t}{t^{2}} ,
		\label{eq:app_fwd_score}
	\end{equation}
	while the marginal score follows from Tweedie's formula, which for this kernel states $\hat x_0^{\old}\equiv\mathbb{E}[x_0|x_t,c]=\big(x_t+t^{2}\nabla_{x_t}\log p^\old_t(x_t|c)\big)/(1-t)$ and hence $\nabla_{x_t}\log p^\old_t(x_t|c)=\big((1-t)\hat x_0^{\old}-x_t\big)/t^{2}$. Subtracting the two,
	\begin{equation}
		\begin{split}
			\nabla_{x_t} \log p^{\old}(x_0 | x_t, c)
			&= \frac{(1-t) x_0 - x_t}{t^2} - \frac{(1-t) \hat{x}_0^{\old} - x_t}{t^2}\\
			&= \frac{1-t}{t}\cdot\frac{x_0 - \hat{x}_0^{\old}}{t}\\
			&= -\frac{1-t}{t} \big( u_t(x_t | x_0) - v^{\old}(x_t, c, t) \big),
			\label{eq:appendix_score_velocity_link}
		\end{split}
	\end{equation}
	where the last equality is just a renaming: $u_t(x_t|x_0)=(x_t-x_0)/t$ is the conditional target and $v^{\old}(x_t,c,t)=(x_t-\hat x_0^{\old})/t$ the reference field, so $(x_0-\hat x_0^{\old})/t = -(u_t-v^\old)$. The posterior score has been traded for a quantity computable from one sample.
	
	\paragraph{Step 4: collection.}
	Chaining \eqref{eq:app_grad_phi}, \eqref{eq:app_logderiv} and \eqref{eq:appendix_score_velocity_link},
	\begin{equation}
		v^{*}-v^{\old}
		= \underbrace{-\frac{t}{1-t}}_{\text{\eqref{eq:app_grad_phi}}}\cdot
		\mathbb{E}\bigg[\frac{w}{z}\cdot\underbrace{\Big(-\frac{1-t}{t}\Big)}_{\text{\eqref{eq:appendix_score_velocity_link}}}\big(u_t-v^\old\big)\bigg]
		= \mathbb{E}_{x_0\sim p^\old(\cdot|x_t,c)}\bigg[\frac{w(c,x_0)}{z(x_t,c,t)}\big(u_t-v^\old\big)\bigg],
		\label{eq:app_collect}
	\end{equation}
	since the two prefactors are reciprocal and the two signs cancel. This is \eqref{eq:theorem_final} exactly. It may be inserted, along with any constant $b$, because $v^\old=\mathbb{E}[u_t|x_t,c]$ forces $\mathbb{E}_{p^\old(\cdot|x_t,c)}[u_t-v^\old]=0$ and hence
	\begin{equation}
		\mathbb{E}\Big[\big(\tfrac{w}{z}-b\big)\big(u_t-v^\old\big)\Big]
		= \mathbb{E}\Big[\tfrac{w}{z}\big(u_t-v^\old\big)\Big] - b\underbrace{\mathbb{E}\big[u_t-v^\old\big]}_{=\,0}
		= v^{*}-v^{\old}, \quad \forall b \in \mathbb{R} .
		\label{eq:app_baseline_free}
	\end{equation}
	Among a plethora of potential choices, $b=1$ is distinguished by $\mathbb{E}_{p^\old(\cdot|x_t,c)}[w/z]=1$, immediately from $\mathbb{E}[w|x_t]=z$, which makes the reweighting zero-mean, while $b=b^*$ of \eqref{eq:optimal_cv} instead minimizes the second moment.
	
	\section{The linear weight implied by the DiffusionNFT branches}
	\label{app:nft_linear_derivation}
	
	This appendix derives \eqref{eq:nft_weight_linear}. DiffusionNFT introduces a binary optimality variable $o$ and identifies the reward with $p(o=1\mid x_0,c)$, so its positive branch is the sampling law conditioned on $o=1$. Bayes' rule at noise level $t$ therefore relates the two noised marginals by the optimality posterior,
	\begin{equation}
		\label{eq:app_bayes_positive}
		\begin{aligned}
		p^{+}_t(x_t\mid c)&=p^{\old}_t(x_t\mid c)\,\frac{p(o=1\mid x_t,c)}{p(o=1\mid c)},\\
		p(o=1\mid x_t,c)&=\mathbb{E}_{x_0\sim p^{\old}(\cdot\mid x_t,c)}\big[R(c,x_0)\big] = \alpha(x_t, c),
		\end{aligned}
	\end{equation}
	where the second identity is the tower property. Comparing \eqref{eq:app_bayes_positive} with the generic tilt of \S\ref{sec:nft_linear_tilt} identifies the branch as a tilt whose weight is the reward itself, up to normalization: 
	\begin{equation}
		w(x_0, c) \propto R(c, x_0),\qquad z(x_t, c, t) \propto \alpha(x_t, c).
	\end{equation}
	
	This is already the substance of the claim, since the weight is affine in the reward rather than exponential in it, but it is worth seeing the velocity-level statement as well. Applying \eqref{eq:value_gradient_velocity} to the pair $(p^+_t,p^\old_t)$ and canceling the $x_t$-independent denominator of \eqref{eq:app_bayes_positive},
	\begin{equation}
		\label{eq:velocity_to_optimality}
		v^{+}(x_t,c,t)-v^{\old}(x_t,c,t)
		= -\frac{t}{1-t}\,\nabla_{x_t}\log\frac{p^{+}_t(x_t\mid c)}{p^{\old}_t(x_t\mid c)}
		= -\frac{t}{1-t}\,\nabla_{x_t}\log\alpha(x_t, c),
	\end{equation}and expanding the remaining gradient exactly as in \cref{app:rectification} (differentiate under the integral, then substitute \eqref{eq:appendix_score_velocity_link}) yields a weight affine in the reward, $v^*=v^\old+\mathbb{E}[(R/\alpha-R_b)(u_t-v^\old)]$, with $R_b$ free through the zero-mean property of the residual. The denominator here is the posterior quantity $\alpha(x_t,c)=\mathbb{E}_{x_0\sim p^\old(\cdot|x_t,c)}[R]$ of \eqref{eq:app_bayes_positive}, which is the $z$ of \eqref{eq:theorem_final} for the weight $w=R$. Replacing it by its endpoint counterpart $\bar R(c)$ is the same substitution that \S\ref{subsec:objective_recovery} makes for the exponential tilt and \cref{app:chi2_derivation} for the projected one, and absorbing the constant into $R_b$ leaves a weight linear in the reward with no posterior quantity in it. The cost is the factor $r_t=\alpha(x_t,c)/\bar R(c)$ proved by \cref{app:z_vs_Z}, a rescaling of the step length that leaves its direction unchanged. Setting $R_b=\bar R(c)$ and scaling by $1/(\beta C\sigma)$ then arrive at \eqref{eq:nft_weight_linear}, the group-centered advantage \eqref{eq:lin_tilt_interior} at temperature $\beta C\sigma$. The implementation and the construction therefore match: the linear tilt is what DiffusionNFT's own branch model implies, and what is chosen without derivation is only the temperature.
	
	\section{Normalization at the endpoint and intermediate level}
	\label{app:z_vs_Z}
	
	Nothing in this section is specific to one tilt: $w$ is an arbitrary non-negative weight, so the statements below apply verbatim to all three members of the family, with $w=\exp(R/\gamma)$ and $Z=Z(c)$ for the exponential tilt, $w=[R-\tau]_+/\gamma$ and $Z=1$ for the projected linear tilt, and $w=R$ with $Z=\bar R(c)$ for DiffusionNFT's dense linear tilt. This is the step the three share, and the only one whose cost has to be argued rather than derived.
	
	Throughout this section, $c$ and $t\in(0,1)$ are fixed and suppressed where possible. Write $q(x_t|x_0)$ for the Gaussian noising kernel of \eqref{eq:unified_target}, $w=w(x_0)$ for the tilt, and
	\begin{equation}
		Z \;=\; \mathbb{E}_{x_0\sim p^\old(\cdot|c)}[w],
		\qquad
		z(x_t) \;=\; \mathbb{E}_{x_0\sim p^\old(\cdot|x_t,c)}[w] .
	\end{equation}
	
	\paragraph{Step 1: the ratio of the two normalizers is the density ratio at level $t$.}
	Marginalizing the tilted endpoint law \eqref{eq:exp_tilt_sol} through the kernel and inserting $p^\old_t(x_t)$,
	\begin{align}
		p^*_t(x_t)
		&= \int q(x_t|x_0)\,p^\old(x_0)\,\frac{w}{Z}\,\mathrm{d}x_0
		\;=\; \frac{p^\old_t(x_t)}{Z}\int \frac{q(x_t|x_0)p^\old(x_0)}{p^\old_t(x_t)}\,w\,\mathrm{d}x_0 \notag \\
		&= \frac{p^\old_t(x_t)}{Z}\,\mathbb{E}_{x_0\sim p^\old(\cdot|x_t)}[w]
		\;=\; p^\old_t(x_t)\,\frac{z(x_t)}{Z} ,
		\label{eq:app_ratio_identity}
	\end{align}
	where the third equality is Bayes' rule. Hence
	\begin{equation}
		r_t(x_t) \;:=\; \frac{z(x_t)}{Z} \;=\; \frac{p^*_t(x_t|c)}{p^\old_t(x_t|c)} \;>\;0 ,
		\label{eq:app_r_def}
	\end{equation}
	which is the quantity whose logarithmic gradient \eqref{eq:theorem_final} is built from. Positivity is immediate since $w>0$.
	
	\paragraph{Step 2: replacing $z$ by $Z$ rescales the displacement.}
	Let $\delta=u_t(x_t|x_0)-v^\old(x_t,c,t)$. We assume throughout this step that $v^\old$ is the exact marginal field of $p^\old_t$, so that
	\begin{equation}
		\mathbb{E}\big[\delta \,\big|\, x_t\big] \;=\; \mathbb{E}_{x_0\sim p^\old(\cdot|x_t,c)}\big[u_t(x_t|x_0)\big] - v^\old(x_t,c,t) \;=\; 0 .
		\label{eq:app_anchor_exact}
	\end{equation}
	
	For any baseline $b$ and any positive $\lambda$ not depending on $x_0$,
	\begin{equation}
		\mathbb{E}\Big[\Big(\tfrac{w}{\lambda}-b\Big)\delta \,\Big|\, x_t\Big]
		= \tfrac{1}{\lambda}\,\mathbb{E}\big[w\,\delta \,\big|\, x_t\big]
		\;-\; b\,\underbrace{\mathbb{E}\big[\delta \,\big|\, x_t\big]}_{=\,0}
		= \tfrac{1}{\lambda}\,\mathbb{E}\big[w\,\delta \,\big|\, x_t\big] ,
		\label{eq:app_lambda_scaling}
	\end{equation}
	so the conditional mean is proportional to $\lambda^{-1}$ and excludes $b$. Taking $\lambda=z(x_t)$ recovers the exact displacement of \eqref{eq:theorem_final}, $v^*-v^\old=z^{-1}\mathbb{E}[w\delta|x_t]$. The implemented target is the bracketed quantity of \eqref{eq:group_loss}, namely $y=v^\old+\hat{\mathcal{A}}\delta$ with $\hat{\mathcal{A}}=w/Z-b$. Conditionally on $x_t$ the anchor $v^\old(x_t,c,t)$ is deterministic and leaves the expectation, so $\mathbb{E}[y|x_t]-v^\old=\mathbb{E}[(w/Z-b)\delta\,|\,x_t]$, and this is \eqref{eq:app_lambda_scaling} evaluated at $\lambda=Z(c)$. Dividing by the same quantity at $\lambda=z(x_t)$,
	\begin{equation}
		\mathbb{E}[y|x_t]-v^\old
		= \frac{1}{Z}\,\mathbb{E}[w\delta|x_t]
		= \frac{z(x_t)}{Z}\cdot\frac{1}{z(x_t)}\,\mathbb{E}[w\delta|x_t]
		= r_t(x_t)\,\big(v^*-v^\old\big) ,
	\end{equation}
	which is \eqref{eq:Z_rescaling}. Two remarks follow immediately. First, the factor $r_t$ is a positive scalar, so the direction of the displacement is preserved exactly and only its length changes. Second, since \eqref{eq:app_lambda_scaling} holds for every $b$, no choice of baseline can repair or worsen the discrepancy. In other words, the baseline controls variance, never this bias.
	
\begin{figure}[t]
		\centering
		\includegraphics[width=\linewidth]{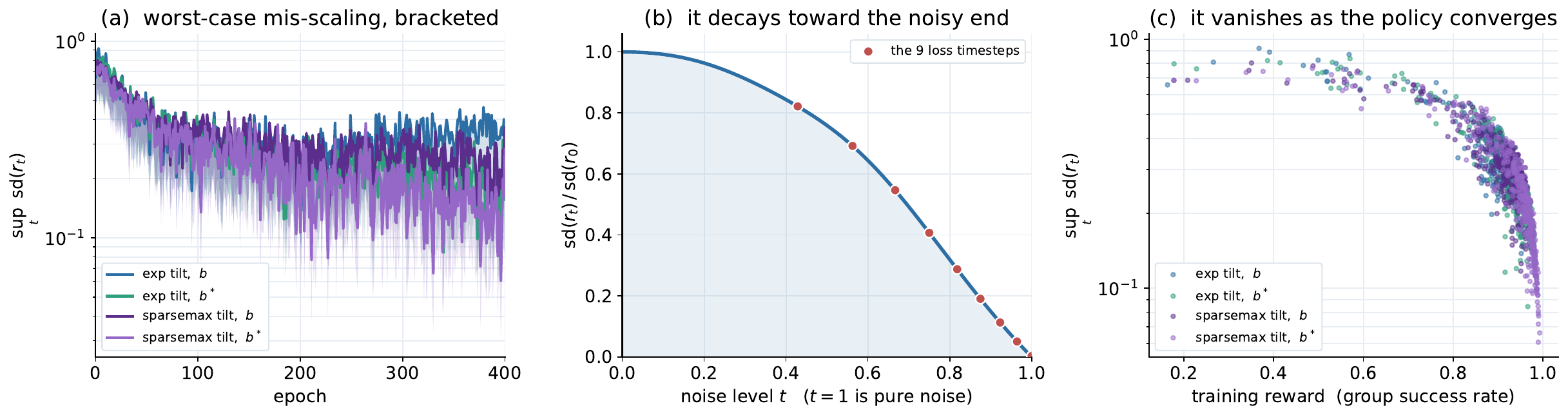}
		\caption{\textbf{How far the implemented target is from the exact one.} By \eqref{eq:app_sup_var} the rescaling $r_t$ of \eqref{eq:Z_rescaling} has mean one at every noise level and variance bounded by $\mathbb{E}[\mathcal{A}^2]$, a property of the reward group alone. \textbf{(a)} That bound, measured on four runs and plotted as a bound on the standard deviation, so every quantity is shown as a square root. No reward model is presumed. For any centered $\mathcal{A}$, $(\mathbb{E}|\mathcal{A}|)^2\le\mathbb{E}[\mathcal{A}^2]\le\max_i|\mathcal{A}_i|\,\mathbb{E}|\mathcal{A}|$. The band is that bracket and the line its upper end. \textbf{(b)} The decay of $\operatorname{sd}(r_t)$ with $t$, computed in a closed form for a Gaussian-mixture endpoint law, which is exact because both exponential tilting and Gaussian smoothing preserve the family. Dots mark the first nine timesteps a $T{=}10$ trajectory-coupled buffer supplies under the $\mathrm{shift}=3$ schedule. When averaged over them, the dispersion is $35\%$ of its worst case. It is a surrogate law and not the model, so this panel is illustrative whereas (a) and (c) are measured. \textbf{(c)} The same bound against the reward it was measured at. As the policy converges the group becomes homogeneous, the advantages shrink and the approximation becomes exact in the limit that matters.}
		\label{fig:misscaling}
	\end{figure}

	Note that in practice, the scope of this statement deserves revising.
	Replacing the posterior mean $z(x_t,c,t)$ by the population endpoint normalizer $Z(c)$ rescales the displacement by a known positive factor, which is the subject of \eqref{eq:Z_rescaling}. The second moment of this factor appears in \eqref{eq:app_sup_var}, and at $t=0$ the posterior degenerates, so $r_0=w/Z$ and $\operatorname{Var}(r_0)=\mathbb{E}[(w/Z-1)^2]$, which is $\mathbb{E}[\mathcal{A}^2]$ at $b=1$.  Replacing $Z(c)$ in turn by the group average $\hat Z=\tfrac1G\sum_j w_j$ leads to \eqref{eq:Softmax_Advantage}, the self-normalized estimator in our implementation. $\hat Z$ is unbiased for $Z$, but $\kappa_i=w_i/\hat Z$ is a ratio in which $\hat Z$ sits in its own denominator, so the conditional expectation does not pass through it and an $\mathcal{O}(1/G)$ bias term remains. At the default setting $G=24$ the bias is not negligible, and we do not claim it is entirely absorbed by the rescaling. The bound is therefore a statement about $Z$, while the diagnostics in \figref{fig:misscaling} report $\hat Z$, and the two differ by the $\mathcal{O}(1/G)$ term above. The $b^*$ of \eqref{eq:optimal_cv} bears a term of the same kind at the smaller micro-batch size.

	\paragraph{Step 3: a logarithmic derivative becomes a plain one.}
	Substituting \eqref{eq:app_r_def} into the score-to-residual identity \eqref{eq:appendix_score_velocity_link} used in \cref{app:rectification} gives the exact target as
	\begin{equation}
		v^*-v^\old \;=\; -\frac{t}{1-t}\,\nabla_{x_t}\log r_t(x_t) ,
		\qquad
		\mathbb{E}[y|x_t]-v^\old \;=\; -\frac{t}{1-t}\,\nabla_{x_t} r_t(x_t) ,
		\label{eq:logr_vs_r}
	\end{equation}
	the second following from the first by multiplying by $r_t$, as Step 2 instructs, since $r_t\nabla\log r_t=\nabla r_t$. The exact update follows the gradient of $\log r_t$, while the implemented one follows the gradient of $r_t$. They agree in direction everywhere and in magnitude whenever $r_t=1$.
	
	\paragraph{Step 4: the dispersion of $r_t$ is monotone in the noise level.}
	Rather than expand near $t=1$, note that the whole path is a smoothing kernel. Dividing $x_t=(1-t)x_0+t\epsilon$ by $(1-t)$ yields
	\begin{equation}
		\frac{x_t}{1-t} \;=\; x_0 + \sigma\epsilon , \qquad \sigma \;=\; \frac{t}{1-t}
		\quad\text{strictly increasing on } (0,1) ,
		\label{eq:app_rescale}
	\end{equation}
	so up to a deterministic, $x_0$-independent rescaling the noising is Gaussian smoothing at level $\sigma$. A deterministic invertible rescaling changes both densities by the same Jacobian and therefore leaves their ratio unchanged, so $r_t$ equals the density ratio of $p^*$ against $p^\old$ after smoothing at level $\sigma(t)$. Gaussian smoothing is a Markov semigroup, for $\sigma_2>\sigma_1$ the law at $\sigma_2$ is obtained from the law at $\sigma_1$ by convolution with $\mathcal{N}(0,(\sigma_2^2-\sigma_1^2)I)$. Two consequences follow. First,
	\begin{equation*}
		\mathbb{E}_{p^\old_t}\big[r_t\big] \;=\; \int \frac{p^*_t}{p^\old_t}\,p^\old_t \;=\; \int p^*_t \;=\; 1\,,\,  \forall\, t
	\end{equation*}
	so the rescaling is centered at one and cannot introduce a systematic inflation or contraction of the step. Second, $\operatorname{Var}_{p^\old_t}(r_t)=\chi^2(p^*_t\|p^\old_t)$, and since the $\chi^2$ divergence is an $f$-divergence with $f(u)=(u-1)^2$, the data-processing inequality applied to the convolution kernel gives
	\begin{equation}
		\chi^2\big(p^*_{t_2}\,\|\,p^\old_{t_2}\big) \;\le\; \chi^2\big(p^*_{t_1}\,\|\,p^\old_{t_1}\big),\, \forall\, t_2 > t_1 ,
		\label{eq:app_dpi}
	\end{equation}
	with supremum at $t\to0$, where the smoothing disappears and the ratio reduces to the endpoint ratio $w/Z$ itself:
	\begin{equation}
		\sup_{t\in(0,1)} \operatorname{Var}_{p^\old_t}\big(r_t\big)
		\;=\; \operatorname{Var}_{x_0\sim p^\old(\cdot|c)}\!\Big(\frac{w}{Z}\Big)
		\;=\; \mathbb{E}\big[\mathcal{A}^2\big] ,
		\label{eq:app_sup_var}
	\end{equation}
	using $\mathcal{A}=w/Z-1$ and $\mathbb{E}[\mathcal{A}]=0$. This is precise and non-asymptotic, and it bounds the discrepancy by a statistic of the reward group that any run already records, which is what \figref{fig:misscaling} exploits. It also localizes the error correctly. The substitution is worst at the clean end of the timestep schedule and asymptotically exact at the noisy end, so a trajectory-coupled buffer, whose timesteps concentrate near $t=1$, samples it where it is smallest.
	
	\paragraph{Consequence for the minimizer.}
	\Cref{app:minimizer} shows that drawing a single $(t,\epsilon)$ pair leaves the minimizer of the regression untouched. That argument applies to whichever conditional mean the target happens to have. Taking \eqref{eq:Z_rescaling} into account, the objective \eqref{eq:group_loss} actually implemented is minimized, over an unconstrained field, at
	\begin{equation}
		v_\theta(x_t,c,t) \;=\; v^\old(x_t,c,t) + r_t(x_t)\,\big(v^*(x_t,c,t)-v^\old(x_t,c,t)\big) ,
		\label{eq:app_impl_minimizer}
	\end{equation}
	not at $v^*$. The gap is a pure step-length modulation and it is not removable by tuning $b$, but \eqref{eq:app_sup_var} bounds its dispersion by $\mathbb{E}[\mathcal{A}^2]$ uniformly in $t$. Two practical readings follow. Where the tilted law already puts more mass than the sampling law, $r_t>1$ and the update is amplified, and where it puts less, the update is damped. And because $r_t$ multiplies the entire displacement, it acts exactly as a state-dependent learning rate, which is why the loss scale $s$ of \eqref{eq:eff_step} and the temperature $\gamma$ are the two most sensitive knobs in \S\ref{sec:ablation}.
	
	\paragraph{Remark.}
	The substitution also has a reading that renders it less like an approximation but more like a choice. Since $p^*(x_0|c)/p^\old(x_0|c)=w/Z(c)$ by \eqref{eq:exp_tilt_sol}, the implemented advantage is $\hat{\mathcal{A}}=p^*(x_0|c)/p^\old(x_0|c)-1$, the importance weight of the tilted endpoint law against the sampling law, centered by one, consistent with the weight of Advantage-Weighted Regression (AWR)~\citep{2019arXiv191000177P}. The substitution is in fact avoidable, yet at another price. Taking logarithms in \eqref{eq:exp_tilt_sol} turns $Z(c)$ into an additive, prompt-only term that any within-group centering removes exactly, which is the route of \citet{2026arXiv260314128Z} and requires estimating the log-ratio of two diffusion models through an ELBO surrogate. Working in the velocity field instead never forms a likelihood ratio, and $r_t$ is what is paid for that. By \eqref{eq:Z_rescaling} the flow-optimal target differs from plain AWR by exactly this factor, which converts the endpoint-level ratio into the level-$t$ ratio.
	
	\section{Why the per-sample target may replace the exact one}
	\label{app:minimizer}
	
	\Cref{eq:theorem_final} expresses the optimal field as a posterior expectation, so the ideal training objective is,
	\begin{equation*}
		\mathcal{L}_{\mathrm{oracle}}(\theta) = \mathbb{E}_{t,\, x_t \sim p_t^\old} \left[ \left\| v_\theta(x_t, c, t) - v^*(x_t, c, t) \right\|^2 \right].
	\end{equation*}
	Evaluating $v^*(x_t,c,t)$ at a single $x_t$ would require averaging over all endpoints consistent with it. The objective actually used replaces that average by one draw,
	\begin{equation}
		\label{eq:proxy_loss}
		\begin{split}
			&\mathcal{L}_{\mathrm{proxy}}(\theta) = \mathbb{E}_{t \sim \mathcal{U}[0,1],\, \epsilon \sim \mathcal{N}(0,I),\, x_0 \sim p^\old(\cdot|c)} \bigg[ \\
			&\quad \Big\| v_\theta(x_t, c, t) - \Big( v^\old(x_t, c, t) + \mathcal{A}(x_0) \cdot \big( u_t(x_t | x_0) - v^\old(x_t, c, t) \big) \Big) \Big\|^2 \bigg],
		\end{split}
	\end{equation}
	which is computable, since every term is a function of the pair $(x_0,x_t)$ at hand.
	
	Substituting a draw for an expectation is legitimate here for the same reason it is legitimate in flow matching itself~\citep{lipman2023flow}. Writing $y(x_0,x_t)$ for the bracketed target of \eqref{eq:proxy_loss} and conditioning on $x_t$,
	\begin{equation}
		\mathbb{E}\big[\|v_\theta - y\|^2 \,\big|\, x_t\big]
		= \big\|v_\theta - \mathbb{E}[y \mid x_t]\big\|^2 + \underbrace{\mathbb{E}\big[\|y - \mathbb{E}[y\mid x_t]\|^2 \,\big|\, x_t\big]}_{\text{independent of }\theta},
		\label{eq:app_bias_variance}
	\end{equation}
	and $\mathbb{E}[y\mid x_t]=v^*(x_t,c,t)$ is by definition in \eqref{eq:theorem_final}. Integrating \eqref{eq:app_bias_variance} over $x_t$ thus yields
	\begin{equation}
		\mathcal{L}_{\mathrm{proxy}}(\theta) \;=\; \mathcal{L}_{\mathrm{oracle}}(\theta) \;+\; C,
		\qquad
		C \;=\; \mathbb{E}_{t,x_t}\Big[\operatorname{Var}\big(y \,\big|\, x_t\big)\Big] \;\perp\; \theta ,
		\label{eq:app_loss_gap}
	\end{equation}
	so the two objectives differ only by a $\theta$-independent constant, their gradients coincide, and minimizing the tractable one drives $v_\theta$ to $v^*$.
	
	This section closes with two concluding remarks. First, the constant $C$ is the conditional variance of the stochastic target, so it does not bias the optimum but does set the variance of the gradient estimator. This is the quantity that the baseline in \eqref{eq:optimal_cv} intends to reduce, and \S\ref{sec:ablation} examines how much that choice matters in practice. Second, because the argument uses only the tower property, it is indifferent to which tilt supplies the reweighting. It applies verbatim to the exponential and linear weightings of \S\ref{app:diffusionnft_kl_equivalence}. The practical consequence is that reward maximization is carried out entirely by supervised regression on reweighted rollouts, with no likelihood ratio, no stochastic sampler, and no differentiation through the reward.
	
	\section{Derivation of the \texorpdfstring{$\chi^2$}{chi-squared} optimum and its group estimator}%
	\label{app:chi2_derivation}
	
	This section completes the three steps that \S\ref{sec:chi2} states without proof: the KKT solution of the $\chi^2$-constrained problem, the intermediate marginals it induces, and the exact group estimator together with its $\bigO(G\log G)$ solution.
	
	\paragraph{Step 1: constrained optimum.}
	Fix a prompt $c$ and write $\rho(x_0)=\frac{\pi_\theta(x_0|c)}{p^\old(x_0|c)}$ as in \eqref{eq:density_ratio}. Problem \eqref{eq:chi2_constrained_objective} reads
	\begin{equation}
		\max_{\rho}\;\; J[\rho] \;=\; \mathbb{E}_{p^\old}\!\left[\rho R\right] - \frac{\gamma}{2}\,\mathbb{E}_{p^\old}\!\left[(\rho-1)^2\right]
		\qquad\text{subject to}\qquad \rho\in\mathcal{K},
		\label{eq:app_chi2_primal}
	\end{equation}
	with $\mathcal{K}$ the feasible set defined in \S\ref{sec:chi2}. The map $\rho\mapsto J[\rho]$ is strictly concave, being the sum of a linear term and the strictly concave $-\frac{\gamma}{2}(\rho-1)^2$, and $\mathcal{K}$ is convex and contains $\rho\equiv1$. The maximizer therefore exists and is unique, and the KKT conditions are necessary and sufficient. Introducing $\lambda\in\mathbb{R}$ for the normalization and $\mu(x_0)\ge0$ for the non-negativity constraint,
	\begin{equation}
		\mathcal{L}[\rho,\lambda,\mu] = \mathbb{E}_{p^\old}\!\left[\rho R - \tfrac{\gamma}{2}(\rho-1)^2\right] - \lambda\Big(\mathbb{E}_{p^\old}[\rho]-1\Big) + \mathbb{E}_{p^\old}\!\left[\mu\rho\right],
		\label{eq:app_lagrangian}
	\end{equation}
	pointwise stationarity $\delta\mathcal{L}/\delta\rho=0$ gives
	\begin{equation}
		R(c,x_0) - \gamma\big(\rho(x_0)-1\big) - \lambda + \mu(x_0) = 0
		\qquad\Longrightarrow\qquad
		\rho(x_0) = 1 + \frac{R(c,x_0)-\lambda+\mu(x_0)}{\gamma}.
		\label{eq:app_stationarity}
	\end{equation}
	Complementary slackness $\mu\rho=0$ with $\mu\ge0$ leaves two possibilities at each $x_0$. Where $\rho>0$ we must have $\mu=0$, so $\rho=1+(R-\lambda)/\gamma$, and this is consistent only if $R>\lambda-\gamma$. Where the constraint binds, $\rho=0$ and $\mu=\lambda-\gamma-R\ge0$, i.e.\ $R\le\lambda-\gamma$. Writing $\tau=\lambda-\gamma$ and combining the branches,
	\begin{equation}
		\rho^\star(x_0) = \left[\,1+\frac{R(c,x_0)-\lambda}{\gamma}\,\right]_+ = \frac{1}{\gamma}\Big[R(c,x_0)-\tau\Big]_+ ,
		\label{eq:app_rho_star}
	\end{equation}
	which is indeed \eqref{eq:lin_tilt_sol}. Imposing $\mathbb{E}_{p^\old}[\rho^\star]=1$ yields \eqref{eq:waterfill} for $\tau$. Uniqueness of its root stems from the fact that $\Phi(\tau)=\mathbb{E}_{p^\old}[(R-\tau)_+]$ is convex (a supremum of affine functions of $\tau$), non-increasing, strictly decreasing wherever $\Pr[R>\tau]>0$, with $\lim_{\tau\to-\infty} \Phi(\tau)\to\infty$ and $\Phi(R_{\max})=0$. If the truncation is inactive, $(R-\tau)_+=R-\tau$ almost surely and \eqref{eq:waterfill} becomes $\bar R-\tau=\gamma$, giving $\tau=\bar R-\gamma$ and hence $\lambda=\bar R$. Substituting into \eqref{eq:app_rho_star} recovers \eqref{eq:lin_tilt_interior}. Inactivity requires $\rho^\star>0$ everywhere, i.e.\ $\operatorname*{ess\,inf}R>\bar R-\gamma$, which is \eqref{eq:validity}.
	
	The group-size bound quoted in \S\ref{sec:chi2} follows from this condition. Writing $z_i=(R_i-\bar R)/\sigma$ for the reward standardized within the group, with $\sigma$ the population standard deviation, $\sum_i z_i=0$ and $\sum_i z_i^2=G$ force $|z_i|\le\sqrt{G-1}$, with equality when one sample differs and the remaining $G-1$ coincide. DiffusionNFT's weight is \eqref{eq:lin_tilt_interior} with $\mathcal{A}_i=z_i/(\beta C)$, so $\min_i\mathcal{A}_i>-1$ is implied by $\beta C\ge\sqrt{G-1}$, that is $G\le1+(\beta C)^2$, and the implication is tight: a one-hot group at $G>1+(\beta C)^2$ violates it. The condition involves only the temperature and the group size, never the rewards.
	The bound is a worst case over reward configurations and the quantifier matters: exceeding it withdraws the guarantee rather than asserting a violation, and whether a particular run is in fact feasible is addressed by monitoring $\min_i\mathcal{A}>-b$, as \S\ref{sec:ablation} does. The default configuration sets $\beta C=5$ and $G=24$, inside that bound with a margin of $4\%$ in the temperature, the ceiling in $G$ being $26$: DiffusionNFT's heuristic keeps itself feasible by construction, but only at its own group size and without saying so, whereas the projection is feasible at every temperature and every $G$.
	
	\paragraph{Step 2: intermediate marginals.}
	The threshold $\tau$ is determined by the terminal marginal $p^\old(\cdot|c)$ and is therefore a constant, not a function of $x_t$. Specializing \eqref{eq:p_t^*} to $w=\rho^\star$ gives
	\begin{equation}
		\begin{split}
			p^*_t(x_t|c) &= p^\old_t(x_t|c)\cdot z(x_t,c,t), \\
			z^\star(x_t,c,t) &= \mathbb{E}_{x_0\sim p^\old(\cdot|x_t,c)}\big[\rho^\star(x_0)\big] = \frac{1}{\gamma}\,\mathbb{E}_{x_0\sim p^\old(\cdot|x_t,c)}\big[(R-\tau)_+\big].
		\end{split}
		\label{eq:app_psi}
	\end{equation}
	No \emph{endpoint} partition function appears, because \eqref{eq:waterfill} already normalizes $\rho^\star$. Here we have $Z=\mathbb{E}_{p^\old(\cdot|c)}[\rho^\star]=1$ exactly, whereas the exponential tilt must estimate its $Z(c)$. The posterior normalizer $z^\star$ is a different matter. The truncation prevents the expectation from passing through the tilt, so $z^\star$ has no closed form in terms of $\bar R(x_t)$, and Step 3 does not compute it at all. Normalizing on the group is normalizing at the endpoint level, so the implemented advantage $\rho_i-b$ is \eqref{eq:theorem_final} with $z^\star$ set to $1$. This is the same substitution \S\ref{subsec:objective_recovery} makes for the exponential tilt, and it is the one step of the construction that every member of the family shares. Its cost is characterized in \cref{app:z_vs_Z} and takes a simpler form here. Since $Z=1$, the factor $r_t=z/Z$ of \eqref{eq:Z_rescaling} is $z^\star$ itself, which is exactly the ratio $p^*_t/p^\old_t$ appearing in \eqref{eq:app_psi}. The direction of the update is therefore preserved and only its length is rescaled, by a factor that is centered at one and tends to one at the noisy end of the schedule. Substituting \eqref{eq:app_psi} into \eqref{eq:value_gradient_velocity} and reusing the score-to-residual identity \eqref{eq:appendix_score_velocity_link}, which is independent of the tilt, yields \eqref{eq:group_loss} with $\mathcal{A}_{\chi^2}=\rho^\star/z^\star-b$, the $\chi^2$ counterpart of the optimal velocity field.
	
	\paragraph{Step 3: the group estimator is a simplex projection.}
	Replace $p^\old(\cdot|c)$ by the empirical measure on a group $\{x_0^{(i)}\}_{i=1}^G$, so each atom carries mass $1/G$, and let $w_i=\rho_i/G$ be the probability assigned by $\pi_\theta$. The constraints $\mathbb{E}[\rho]=1$ and $\rho\ge0$ become $\sum_i w_i=1$ and $w_i\ge0$, that is $w\in\Delta^{G-1}$, and \eqref{eq:app_chi2_primal} becomes
	\begin{equation}
		\max_{w\in\Delta^{G-1}}\;\; \langle w, R\rangle - \frac{\gamma G}{2}\left\|w-\tfrac{1}{G}\mathbf{1}\right\|^2
		\;\;\Longleftrightarrow\;\;
		\min_{w\in\Delta^{G-1}}\;\; \left\|w - \left(\tfrac{1}{G}\mathbf{1}+\tfrac{R}{\gamma G}\right)\right\|^2 ,
		\label{eq:app_projection}
	\end{equation}
	where the equivalence follows by expanding the square and discarding the terms independent of $w$. The right-hand problem is by definition the Euclidean projection onto the simplex, i.e.\ $\operatorname{sparsemax}$ operator of~\citep{pmlr-v48-martins16}, which establishes \eqref{eq:sparsemax_adv}. Its solution is obtained by sorting~\citep{10.1145/1390156.1390191}: let $R_{(1)}\ge\dots\ge R_{(G)}$ and
	\begin{equation}
		\tau_k = \frac{1}{k}\left(\sum_{j\le k}R_{(j)} - \gamma G\right),
		\qquad
		k^\star = \max\big\{\,k : R_{(k)} > \tau_k \,\big\},
		\qquad
		\tau = \tau_{k^\star},
		\label{eq:app_sort_solution}
	\end{equation}
	with $k^\star$ being the support size. The set in \eqref{eq:app_sort_solution} is never empty, since $R_{(1)}>\tau_1=R_{(1)}-\gamma G$ for any $\gamma>0$, so $\tau$ is always defined. The cost is $\bigO(G\log G)$, dominated by the sort. Finally, because $\tau$ is chosen so that $\sum_i\rho_i=G$, the empirical partition satisfies $\hat{z}=\tfrac1G\sum_i\rho_i=1$ exactly, which is why $z$ need not be estimated. Note that this identity holds by construction of $\tau$, and not, as in the dense case, because the centered weights happen to average to one.
	
	\section{Implementation and experimental details}
	\label{app:impl_details}
	
	Every run in \S\ref{sec:design_space} adopts the following setup inherited from DiffusionNFT, although it may place our methods at a disadvantage.
	
	\paragraph{Backbone and adapters.}
	SD3.5-Medium at $512^2$ pix resolution, loaded in \texttt{fp16} mixed precision. Only LoRA~\citep{hu2022lora} adapters are trained, at rank $r=32$ with scaling $\alpha=64$, while the backbone is frozen. This makes the frozen-anchor variants of \S\ref{sec:ablation} inexpensive, since the pretrained field $\vf_{\text{ref}}$ is recovered by disabling the adapters, so no second copy of the weights is held in memory.
	
	\paragraph{Prompt pool and evaluation set.}
	Both splits are the GenEval metadata shipped with the DiffusionNFT codebase. The training pool holds $50{,}000$ records spanning $33{,}199$ distinct prompts and the evaluation split holds $2{,}212$ records over the $553$ distinct prompts of the GenEval benchmark, four records per prompt, and is evaluated in full at every evaluation round rather than subsampled.
	
	\paragraph{Rollout and buffer.}
	An epoch draws $48$ prompts and samples each $G=24$ times, for $1{,}152$ images per epoch. The prompt count follows from the sampler rather than being set directly: with $8$ devices, a per-device batch of $9$ and $G=24$ images per prompt, each batch covers $8\times9/24=3$ prompts, and an epoch runs $16$ such batches. Sampling uses a $T=10$ step first-order Euler solver on the flow-matching schedule with shift $3.0$, and is CFG-free~\citep{ho2021classifierfree}. Each buffer is consumed by a single optimization iteration before the next rollout, so no endpoint is revisited and the frozen-buffer premise of \S\ref{subsec:method_prelims} holds. By default the loss is evaluated at the timesteps the rollout itself visited, which is the coupling that axis \textbf{(iv)} isolates. A fraction $90\%$ of the schedule is retained, which merely excludes the degenerated endpoint.
	
	\paragraph{Optimization.}
	AdamW~\citep{loshchilov2018decoupled} with $\beta=(0.9,0.999)$, $\epsilon=10^{-8}$ and weight decay $10^{-4}$ is utilized, at a constant learning rate of $3\times10^{-4}$. The micro-batch size is $9$ latents per device with $16$ gradient-accumulation steps. Gradients are clipped to unit norm and an exponential moving average of the adapters is maintained. Both are kept fixed across all runs, so the gradient norms reported in \S\ref{sec:ablation} are the pre-clipping values and remain comparable between configurations. The loss scale is $s=5.0$, which is also the $s$ entering $\eta_{\text{eff}}$ in \eqref{eq:eff_step}. For the exponential tilt, $\gamma$ is set adaptively to the reward standard deviation pooled across the batch and floored at $0.01$, and no advantage clipping is applied unless an ablation asks for it. All runs use seed $42$ for reproducibility.
	
	\paragraph{Evaluation.}
	Evaluation runs every $10$ epochs with a $40$-step first-order solver, four times the rollout budget, so that a policy cannot appear to improve merely by adapting to a coarse sampler. No run is stopped early on a training episode. Each is trained until it either plateaus or collapses, within a maximum budget of $1{,}000$ epochs, and the horizon shown in a figure is the horizon that was run.
	
	\paragraph{Diagnosis metrics.} We also fix three quantities that we record for every run and that turn out to explain most of what we observe in the incoming section. The first is the \emph{effective step size}
	\begin{equation}
		\eta_{\text{eff}} \;=\; s\cdot\mathbb{E}\big|\mathcal{A}\big| ,
		\label{eq:eff_step}
	\end{equation}
	where $s$ is the loss scale of \eqref{eq:unified_target}. Since the regression target departs from the anchor by exactly $\mathcal{A}\,\delta$, the quantity $\eta_{\text{eff}}$ measures, up to the residual norm, how far each update asks the policy to move; it is the natural common currency for comparing an exponential and a linear tilt, whose advantages have different scales by construction. The second is $\|\vf_\theta-\vf_\old\|^2$. Here $\vf_\old$ is the exponential moving average of the policy, refreshed once per epoch and maintained in every configuration, so this is the policy--EMA gap: a proxy for how much the policy is currently changing. When the anchor is frozen it is in particular not the distance to the anchor, which is reported separately as $\|\vf_\theta-\vf_{\text{ref}}\|^2$, the squared distance to the adapter-disabled pretrained field averaged over the batch and over the sampled $t$ with the weights the loss itself uses. We report that second quantity as an implicit KL because it is proportional to the per-step KL under the Gaussian transitions, and all read-outs of both are medians over an explicitly stated epoch window. The third is the fraction of prompt groups whose reward has zero variance, which measures how much of the batch still carries a usable learning signal as the reward saturates.
	
	\begin{figure}[t]
		\begin{center}
			\includegraphics[width=\linewidth]{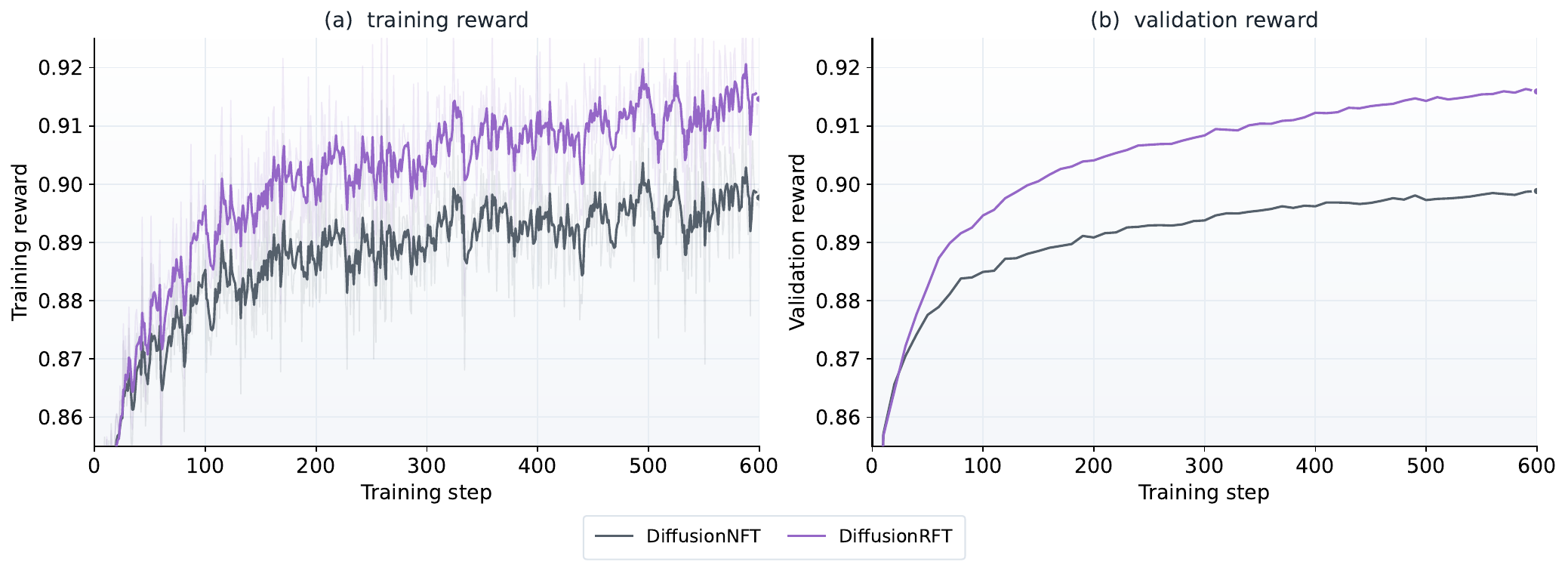}
		\end{center}
		\caption{\textbf{Generalization to Pickscore reward.} DiffusionRFT achieves consistent superiority over DiffusionNFT, as displayed by training reward (a) and validation reward (b). }
		\label{fig:pickscore-gen}
	\end{figure}
	
	\paragraph{Generalization experiments.}
	To substantiate the generality and scalability of our method, we replicated the experiments on Pickscore, a learned preference model instead of a rule-based reward. As demonstrated in~\figref{fig:pickscore-gen}, the results are consistent with the main paper. Our method delivers markedly faster convergence and reaches a higher performance ceiling than the competitor.
	
	\section{Per-run read-outs for the ablations}
	\label{app:ablation_details}
	
	\subsection{The prediction space and the irreducible residual}
	\label{app:pred_space}
	
	This section explains \figref{fig:loss-dynamics}: why the same update, written in $\vv$ rather than in $\vx$, leaves an order of magnitude more regression loss at the same movement budget.
	
	\paragraph{The residual carries no reward.} Write $\Delta=v_\theta-v^\old$ for the displacement the network realizes and $\delta=u_t-v^\old$ for the per-sample residual, so that the objective is $L=\mathbb{E}\Vert\Delta-\mathcal{A}\delta\Vert^2$. Minimizing over $\Delta$ at fixed $x_t$ gives $\Delta^\star=\mathbb{E}[\mathcal{A}\delta|x_t]$ and
	\begin{equation}
		L^\star \;=\; \mathbb{E}\big[\mathcal{A}^2\Vert\delta\Vert^2\big] \;-\; \big\Vert\mathbb{E}[\mathcal{A}\delta\,|\,x_t]\big\Vert^2 .
		\label{eq:app_irreducible}
	\end{equation}
	This is a bias--variance decomposition, in which the two terms answer different questions. The regression target $\mathcal{A}\delta$ is random even after $x_t$ is fixed: it depends on which endpoint $x_0$ was drawn and on the reward that endpoint received, neither of which $v_\theta$ can see. A function of $x_t$ alone can therefore match the conditional mean of the target and nothing better, and the subtracted term $\Vert\Delta^\star\Vert^2$ measures how much of the target's size that mean accounts for. Therefore, $\mathbb{E}\Vert\mathcal{A}\delta-\Delta^\star\Vert^2$ is the conditional variance of the target: the spread of the individual $\mathcal{A}\delta$ around the direction the network moves in. No parameterization and no amount of training removes it, so it sets the floor that a loss curve descends towards, and comparing two runs by their loss is comparing their floors.
	
	The first term of \eqref{eq:app_irreducible} factorizes when $\mathcal{A}^2$ and $q=\Vert\delta\Vert^2$ are close to uncorrelated. This is a second-order condition and it does not follow from the small optimal baseline, which constrains $\mathbb{E}[\mathcal{A}q]$ and hence $\operatorname{Cov}(\mathcal{A},q)$ instead in \S\ref{app:baseline}. Probing the within-batch correlation of $\mathcal{A}^2$ against $q$ over a $50$-epoch run gives a median of $0.013$, within a range of $[-0.086,0.097]$, and no drift as the reward saturates, the correlation against the reward being $-0.105$. Note that each logged value aggregates that epoch's micro-batches, so its spread of $0.040$ reflects an effective sample size of several hundred rather than a group of $24$. Writing $\mathbb{E}[\mathcal{A}^2q]=\mathbb{E}[\mathcal{A}^2]\,\mathbb{E}[q]\,(1+\varrho\,\mathrm{cv}(\mathcal{A}^2)\,\mathrm{cv}(q))$ with $\varrho$ being that correlation, the error of the product is set by $\varrho$ together with the two dispersions. We monitor $\mathbb{E}[\mathcal{A}^2q]/(\mathbb{E}[\mathcal{A}^2]\mathbb{E}[q])-1$ per batch over $50$ steps. Its magnitude has median $2.8\%$, reaches $9\%$ at the $95$th percentile, and carries a small positive bias, with the signed mean being $+0.016$ and a standard error of $0.007$. Then
	\begin{equation}
		L^\star \;\approx\; \underbrace{\mathbb{E}\big[\mathcal{A}^2\big]}_{\text{a property of the reward group}}\;\cdot\;\underbrace{\mathbb{E}\Vert\delta\Vert^2}_{\text{a property of the noising kernel}} ,
		\label{eq:app_residual_split}
	\end{equation}
	It approximates $L^\star$ when the correlation above is small, which is measurable. \Figref{fig:loss-dynamics}(d) plots the regression loss with the loss scale divided out rather than a floor extracted from it. What the scale does say is that the advantage enters only through $\mathbb{E}[\mathcal{A}^2]$, so with the advantage held comparable across runs the second factor is what remains varied, which is the sense in which that panel is labeled reward-independent.

	\paragraph{The prediction space sets that factor.} By the affine mapping, $\delta=-\big(x_0-\hat x_0^{\old}\big)/t$, so
	\begin{equation}
		\mathbb{E}\big[\Vert\delta\Vert^2\,\big|\,x_t\big] \;=\; \frac{\operatorname{Var}\big(x_0\,\big|\,x_t\big)}{t^{2}} .
		\label{eq:app_delta_t2}
	\end{equation}
	The numerator is the posterior spread of the endpoint given the noised state, an irreducible quantity of the data and the kernel. For a fixed rollout distribution, this posterior spread cannot be removed by fitting the conditional mean more accurately, and at the noisy end of the schedule it is of the order of the data variance itself. The denominator is decided by the prediction space. Regressing in $\vx$ means comparing $\hat x_0=x_t-t\,v$ against $x_t-t\,v_{\text{target}}$, so the residual acquires a factor $t$ and the $t^{-2}$ of \eqref{eq:app_delta_t2} cancels exactly. The $\vx$ and $\vv$ objectives share a minimizer and differ only in the weight they afford each noise level. In $\vv$ the factor stands out, and over the nine timesteps the schedule retains, $t\in[0.429,1]$, the mean of $t^{-2}$ is $2.08$.
	
	\paragraph{Why this is a misdirection and not a short step.} 
	A large regression loss admits two understandings, and they call for opposite remedies. Either the network has not yet moved far enough, in which case the loss is an excess over the floor and more optimization removes it, or it has moved as far as the target allows and the loss is the floor itself, in which case more optimization changes nothing and the target has to be reformulated. \Eqref{eq:app_irreducible} exactly discriminates these two cases, the excess being what $\Delta$ can still absorb and the floor being what it cannot, so the two understandings are distinguished by whether the movement budget has been spent. \Figref{fig:loss-dynamics} illustrates that the budget has been spent. The $\vv$-space run moves as far as the $\vx$-space one, $0.190$ against $0.177$ in $\|\vf_\theta-\vf_{\text{ref}}\|^2$, and it is handed more signal to move with $\mathbb{E}|\mathcal{A}|=0.538$ against $0.420$. Nonetheless, it ends with $8.9\times$ the residual and less reward, $0.442$ against $0.693$. A run that has spent its budget and still sits high above its counterpart is not taking short steps.
	
	That said, $8.9\times$ is not itself a pure measure of how much worse the fit is, because by \eqref{eq:app_residual_split} the two runs do not share a floor. The floor is a product of two factors and both differ. The advantage factor is $\mathbb{E}[\mathcal{A}^2]_{\vv}/\mathbb{E}[\mathcal{A}^2]_{\vx}$, and we do not log the second moment directly. Each run supplies $L\le\mathbb{E}[\mathcal{A}^2]\le U$ with $L=(\mathbb{E}|\mathcal{A}|)^2$ and $U=\max_i|\mathcal{A}_i|\,\mathbb{E}|\mathcal{A}|$, and a ratio of two bracketed quantities is bracketed only across sides, $L_{\vv}/U_{\vx}\le\mathbb{E}[\mathcal{A}^2]_{\vv}/\mathbb{E}[\mathcal{A}^2]_{\vx}\le U_{\vv}/L_{\vx}$, which at epoch $39$ gives $[0.69,4.07]$. The kernel factor is the $t^{-2}$ of \eqref{eq:app_delta_t2}, worth $2.08$ over the retained timesteps, so the floor ratio is known only to lie in $[1.45,8.47]$, and the measured $8.9$ nearly falls inside it. The excess over the floor is therefore not identified, and these statistics facilitate a scale analysis rather than a decomposition. That is enough for the conclusion it was meant to support: a regression loss is not comparable across prediction spaces, since here the change of floor alone can account for the entire measured ratio.
	
	One of the two floor factors deserves a comment. The advantage factor is larger for the $\vv$-space run, but that is downstream of the failure rather than upstream. Its reward is the lower, $0.442$ against $0.693$, so its groups are more heterogeneous and its advantages the larger. The ratio is $1.2$ over the first $15$ epochs and $3.3$ over epochs $30$ to $45$. A gap that widens as the reward gap widens is consistent with tracking the outcome rather than causing it, but the ordering alone does not establish that. The $t^{-2}$ factor provides a plausible mechanism for the observed gap, and it acts twice: once explicitly, by weighting the floor, and once implicitly, via the weaker policy that weighting generates.
	
	\subsection{Why the dense linear tilt collapses.}
	\label{app:tilt}
	The instability of the unregularized linear tilt is not, however, a property of the linear family as such, and \S\ref{sec:chi2} tells us where to look. Criterion \eqref{eq:validity} states that the centered form \eqref{eq:lin_tilt_interior} is the $\chi^2$ optimum only while every relative tilt stays positive, equivalently $\min_i\mathcal{A}>-b$. At the $b=1$ case used here the quantity to monitor is simply the smallest advantage in the batch. Doing so is decisive: the centered linear tilt spends $86.8\%$ of its training steps outside that regime and reaches $\min_i\mathcal{A}=-2.46$, whereas over $928$ steps of exponential-tilt training the bound $\mathcal{A}\ge-1$ is never approached, as the barrier property of $f_{\text{KL}}$ requires. Among the three runs the terminal drop is monotone in the effective step size \eqref{eq:eff_step}, $\eta_{\text{eff}}=0.71$, $0.90$ and $2.11$ giving drops of $0.001$, $0.046$ and $0.185$. We posit that the collapse is the signature of a violated density constraint, and \eqref{eq:sparsemax_adv} supplies the corresponding repair: retain the projection in place of the dense approximation. We therefore probe the projection-only variant where the sparsemax advantage replaces the centered one, holding the \emph{constant} baseline fixed so that the projection is the only variable that changes. Retaining the projection restores stability on its own: the sparsemax tilt ends $0.001$ below its peak against $0.185$ for the dense linear tilt.

	\subsection{How the optimal baseline comes into play.}
	\label{app:baseline}
	
	Two remarks are important regarding the constant $b$ in \eqref{eq:advantage_base}, which the projection leaves free and which \eqref{eq:optimal_cv} then fixes by minimizing variance.  First, $b=1$ is the centering choice for \emph{every} tilt considered here, and not a convention that happens to suit one of them: the relative tilt $\kappa$ has group mean exactly one for the exponential weight of \eqref{eq:Softmax_Advantage}, for the sparsemax weight of \eqref{eq:sparsemax_adv} by the construction of $\tau$, and for the dense linear weight, whose advantage $(R-\bar R)/\gamma$ in \eqref{eq:lin_tilt_interior} is precisely $\kappa-1$ once $\kappa=1+(R-\bar R)/\gamma$ is recognized.  What varies between tilts is therefore not the centering constant but the spread of $\kappa$ around it.  Second, the baseline shift and the simplex projection of \S\ref{sec:chi2} act on different objects and compose without interference: the projection enforces $\rho\ge0$ and fixes the feasible set, while $b^*$ is a variance-reduction displacement applied afterwards, unbiased for any constant $b$ because $\mathbb{E}[u_t-v^\old]=0$.  Note finally that $b^*$ is estimated on the same micro-batch that it reweights, so the self-normalised control variate involves an $\bigO(1/B)$ finite-batch bias. With the small micro-batches used here this term is not negligible, and the empirically observed sign of $b^*$ corresponds to a slight systematic pull of the regression target toward the sampled data.
	
	\paragraph{How much variance the control variate removes.}
	The baseline of \eqref{eq:advantage_base} leaves a constant free, and \eqref{eq:optimal_cv} spends it on variance: $b^*$ is the additional shift that minimizes the second moment of the regression target. How much it removes has an exact answer. Write $q=\Vert\delta\Vert^2$, and recall that $\mathcal{A}$ is already centered, $\tfrac1G\sum_i\mathcal{A}_i=0$. The quantity \eqref{eq:optimal_cv} minimizes is then $M(b)=\mathbb{E}\big[(\mathcal{A}-b)^2q\big]$, a quadratic in $b$ whose vertex is $b^*=\mathbb{E}[\mathcal{A}q]/\mathbb{E}[q]$, which is \eqref{eq:optimal_cv} itself. Since $\mathcal{A}$ is centered this is exactly $\operatorname{Cov}(\mathcal{A},q)/\mathbb{E}[q]$. Evaluating $M$ at the vertex,
	\begin{equation}
		M(0)-M(b^*) \;=\; \frac{\operatorname{Cov}(\mathcal{A},q)^2}{\mathbb{E}[q]} \;=\; (b^*)^2\,\mathbb{E}[q] ,		\qquad
		\frac{M(0)-M(b^*)}{M(0)} \;=\; \frac{\operatorname{Cov}(\mathcal{A},q)^2}{\mathbb{E}[q]\;\mathbb{E}\big[\mathcal{A}^2q\big]} .
		\label{eq:app_cv_reduction}
	\end{equation}
	Two features of \eqref{eq:app_cv_reduction} matter later. The reduction is governed by the covariance between the advantage and the squared residual, so it vanishes exactly when the two are uncorrelated, however large either of them is. And $M(0)=\mathbb{E}[\mathcal{A}^2q]$ is the second moment already at the centering baseline, so the ratio says what $b^*$ adds to $b=1$ and not what a baseline achieves from scratch.
	
	Putting numbers to it, we notice $|b^*|/\mathbb{E}|\mathcal{A}|\approx1\%$ in practice. When $\mathcal{A}^2$ and $q$ are close to uncorrelated as stated above, the denominator factorizes as $\mathbb{E}[\mathcal{A}^2]\,\mathbb{E}[q]$, so the relative reduction is about $(b^*)^2/\mathbb{E}[\mathcal{A}^2]$, at most $(|b^*|/\mathbb{E}|\mathcal{A}|)^2=10^{-4}$. The reward contribution is correspondingly small, $+0.011\pm0.002$ at a matched budget. By the first feature above, $b^*$ is a covariance divided by $\mathbb{E}[q]$, so a small $b^*$ says that the advantage and the squared residual are nearly uncorrelated in these batches. A parameter that removes $10^{-4}$ of the second moment cannot be what keeps the dense linear tilt from collapsing, and the next subsection asks what it does.
	
	\paragraph{What the control variate does.}
	The grid above is unambiguous on the effect: without a control variate the dense linear tilt ends at $0.718$, with one it ends at $0.965$. Since variance reduction of order $10^{-4}$ cannot account for that, the stabilization must come from some other property of $b^*$, and there are two candidates. It is a random variable, re-estimated on every micro-batch, so it has both a typical size and a tail; either could be doing the work. We separated them with four variants of the dense linear cell, each a single-key change.
	
	Two of the four keep the typical size and remove the tail: one replaces $b^*$ by the constant equal to its logged median, $-1.5\times10^{-3}$, and the other clips $|b^*|$ at a running $90$th percentile. One keeps only the tail, zeroing $b^*$ below that quantile. The fourth caps $|b^*|$ at $1\%$ of the batch advantage magnitude. At a matched $390$ epochs their terminal evaluation rewards are $0.941$, $0.111$, $0.847$ and $0.000$, against $0.718$ with no control variate and $0.964$ with the optimal one. The constant settles the question. It carries no tail, no per-batch adaptivity and no variation of any kind, and it recovers $90\%$ of the gap.
	
	What the set does establish is a statement about the configuration rather than about $b^*$. None of these runs restores feasibility, with $\min_i\mathcal{A}$ staying between $-1.06$ and $-1.32$ throughout so that \eqref{eq:validity} is violated in all of them, and yet a perturbation worth $0.36\%$ of the advantage scale decides whether training ends at $0.718$ or at $0.941$. A configuration on which so small a perturbation is decisive is one at a stability boundary, and single-seed runs at such a boundary cannot separate a mechanism from sensitivity to the perturbation itself. That sensitivity is what the projection removes without a control variate at all: the projected tilt holds its peak to within $0.001$.
	
	\subsection{What confounds anchor-lock with RAM's escape.}
	\label{app:anchor}
	The lock is not a property of the frozen anchor alone, and separating what it depends on also explains why RAM itself does not appear to suffer from the lock. Two quantities are confounded in a faithful RAM configuration. The first is how far the target is permitted to move off the anchor. RAM multiplies its advantage by $100$ before adding it to the frozen base velocity. At unit multiplier the correction is roughly $40\times$ too weak to displace $\vf_\theta$ from $\vf_{\text{ref}}$ at all, which is exactly the regime the three stagnated variants sit in, and exactly the $\eta_{\text{eff}}$ of \eqref{eq:eff_step} being too small for the displacement the anchor demands. The second is where the anchor sits. RAM trains with classifier-free guidance at scale $2.0$, whereas our grid is CFG-free throughout. The base model's own training reward is $0.164$ without guidance against $0.410$ with it, so the two setups anchor to pretrained behaviors of very different quality. \citet{2026arXiv260314128Z} identify precisely the second effect, observing that anchoring to the weaker non-CFG conditional model can dominate the reward term and pull the policy towards low-quality regions, and propose keeping the sampler CFG-free while anchoring the regularizer to the CFG-guided model. RAM does not make that separation: it simply guides both, but still enjoys the benefit.
	
	Raising the multiplier does lift the ceiling. Our faithful port reaches $0.711$ from its higher starting point, and the official implementation reaches $1.000$ in training reward. However, neither survives. Both collapse to zero afterwards, and the diagnostics of \figref{fig:anchor-dynamics} offer an explanation. With the anchor fixed, the displacement the target demands of a single update grows as the policy improves, while $\gamma$ contracts towards its floor as the reward saturates, so the same multiplier that was barely sufficient early turns to be far too large late. In the sequel, we shed further light on the lock phenomenon of frozen anchors.
	
	We concentrate on the signal first. Frozen anchors give $\mathbb{E}|\mathcal{A}|=0.49$ to $0.58$ against $0.29$ for the rolling anchor, and their reward has not saturated. The zero-variance group fraction stays at $0.000$ and the within-group reward standard deviation at $0.40$, whereas under the rolling anchor the first rises from $0$ to $0.479$ and the second falls from $0.485$ to $0.094$. The configuration that wins is the one running out of signal, and the configuration that keeps its signal is the one that stalls.
	
	Two further quantities separate local change from net displacement. The first is the distance to $\vf_\old$. The frozen speed stays within $2.6$ to $3.7\times10^{-6}$ in all three runs, while the rolling speed crosses that ceiling at epoch $8$ and ends four orders of magnitude above it. The second is the distance to $\vf_{\text{ref}}$, which is the quantity we report as an implicit KL and to which it is proportional under the Gaussian transitions. It separates the two configurations with no crossover at all, $1.9$ to $4.5\times10^{-4}$ for the frozen runs against $1.5\times10^{-1}$, and split by fifths of the window it grows $4.4\times$ under the rolling anchor and not at all under a frozen one, ending at $0.3$ to $0.8$ of where it started. The frozen policy is held in a neighborhood, it is not drifting slowly out of one.
	
	Terminal training rewards for the three frozen variants are $0.249$, $0.246$ and $0.248$, against $0.980$ for the rolling-anchor run. For the two RAM implementations, comparing medians over the $80$ epochs before the evaluation peak with the $120$ after it, we find that the official run moves from $0.98$ to $0.92$ in reward and from $0.23$ to $0.36$ in advantage magnitude, while its output-delta norm, a norm rather than the squared distance above, grows $7.2\times$ ($0.031\to0.224$), its target norm from $836$ to $1437$, and its gradient norm $3.4\times$ ($0.46\to1.55$). Our reproduction logs the gradient norm but not the displacement, and shows the same runaway in milder form, $1.3\times$ ($0.57\to0.76$). The failure is therefore a runaway of the target rather than a loss of signal. The anchor is fixed while $\mathcal{A}\,\delta$ is scaled by $100$, the target separates from the field the sampler uses, and the regression amplifies its own error. A rolling anchor removes this feedback path, because the anchor follows the policy and the target cannot separate from the sampling distribution indefinitely.

	\subsection{Why the decoupled buffer is worse.}
	\label{app:buffer}
\begin{figure}[t]
		\begin{center}
			\includegraphics[width=\linewidth]{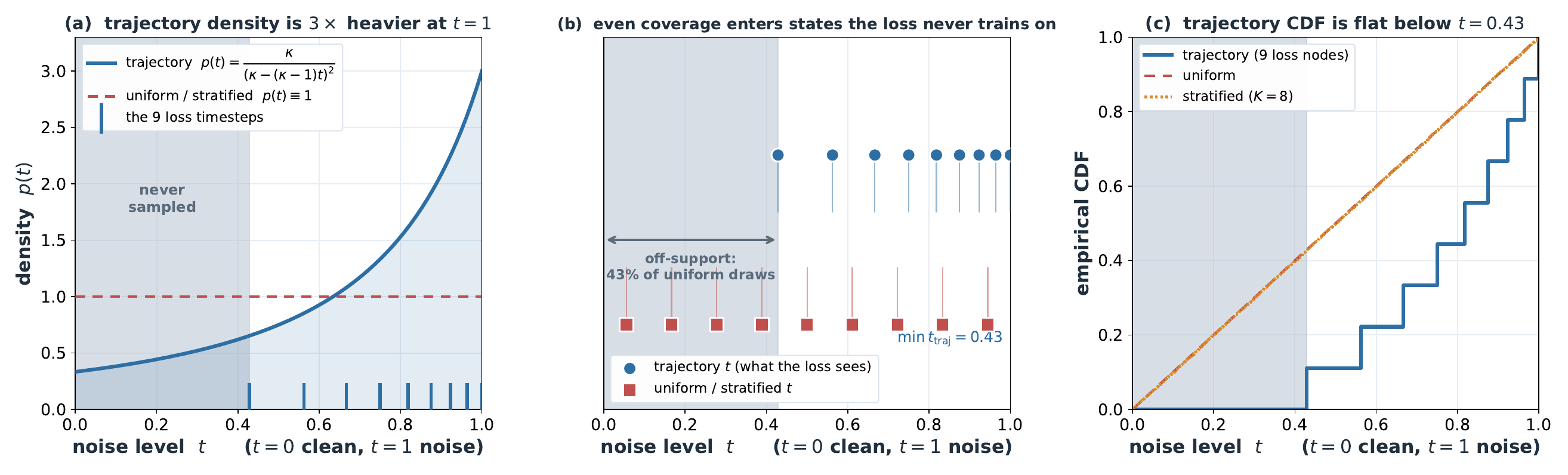}
		\end{center}
		\caption{\textbf{The trajectory schedule covers only part of the noise axis.} $t$ is the noise level, $t=0$ being clean data. (a) Pushing the uniform grid through the SD3.5 shift $\kappa=3$ gives a density $\kappa^2=9\times$ heavier at $t=1$ than at $t=0$, and the nine loss timesteps inherit it. (b) Those nine nodes never fall below $t=0.43$, so the shaded band is off-support for the \emph{loss}: drawing $t\sim\mathcal{U}[0,1]$ sends $43\%$ of the loss evaluations there, and stratification guarantees it. (c) The same statement as a CDF. This is the mechanism behind \figref{fig:buffer-dynamics}: decoupling the renoising timestep does not merely re-weight the noise axis, it regresses at noise levels the policy is never asked to act on.}
		\label{fig:traj-t}
	\end{figure}

	The diagnostics in~\figref{fig:buffer-dynamics} are clearly demarcated: both split by the law of $t$ and are almost invariant in $K$. $\mathbb{E}|\mathcal{A}|$ is $0.118$ and $0.105$ under the trajectory law against $0.237$ and $0.250$ under the uniform one, and the zero-variance group fraction is $0.60$ and $0.62$ against $0.34$ and $0.32$. The within-batch spread of the displacement, as the ratio of batch maximum to batch median, is $1.3\times10^3$ and $3.0\times10^3$ against $2.1\times10^3$ and $5.1\times10^3$, rising to $1.1\times10^4$ once the uniform draws are stratified. What decoupling changes is not the variance of $t$ but its \emph{support}. The sampler walks the $10$-step shifted grid of  $t\in\{0.43,\dots,1.00\}$ in~\figref{fig:traj-t}, so the trajectory schedule never regresses below $t=0.43$, whereas $t\sim\mathcal{U}[0,1]$ places $43\%$ of the loss evaluations in a near-clean band the policy scarcely visits when sampling.
	
\end{document}